\documentclass[final]{vpd}

\usepackage{times}
\usepackage{epsfig}
\usepackage{epstopdf}
\usepackage{graphicx}
\usepackage{amsmath}
\usepackage{amssymb}

\usepackage{subcaption}
\usepackage{tikz}
\usepackage{pgfplots}
\usepackage{enumitem}
\usepackage{booktabs}
\usepackage{multirow}
\usepackage{algorithm}
\usepackage{algpseudocode}
\usepackage{color}

\newcommand{\rect}[2]{-- ++(#1/2, #1/2) -- ++(0, #2) -- ++(-#1/2, -#1/2) -- ++(0, -#2)}

\usepackage[pagebackref=true,breaklinks=true,colorlinks,bookmarks=false]{hyperref}

\def\cvprPaperID{11502} 
\def\confYear{CVPR 2021}

\begin{document}

\title{Variational Pedestrian Detection}

\author{
    Yuang Zhang\textsuperscript{1$\ast$},\
    Huanyu He\textsuperscript{1$\ast$},\
    Jianguo Li\textsuperscript{2},\
    Yuxi Li\textsuperscript{1},\
    John See\textsuperscript{3},\
    Weiyao Lin\textsuperscript{1$\dag$} \\
    \textsuperscript{1}Shanghai Jiao Tong University, China,
    \textsuperscript{2}Ant Group,
    \textsuperscript{3}Heriot-Watt University, Malaysia\\
    }
\maketitle
\thispagestyle{empty}
\pagestyle{empty}
\renewcommand{\thefootnote}{\fnsymbol{footnote}}
\footnotetext[1]{Equally-contributed first authors}
\footnotetext[2]{Corresponding author, Email:wylin@sjtu.edu.cn}

\begin{abstract}
  Pedestrian detection in a crowd is a challenging task due to a high number of mutually-occluding human instances, which brings ambiguity and optimization difficulties to the current IoU-based ground truth assignment procedure in classical object detection methods.
  In this paper, we develop a unique perspective of pedestrian detection as a variational inference problem. We formulate a novel and efficient algorithm for pedestrian detection by modeling the dense proposals as a latent variable while proposing a customized Auto-Encoding Variational Bayes (AEVB) algorithm. Through the optimization of our proposed algorithm, a classical detector can be fashioned into a variational pedestrian detector.
  %
  Experiments conducted on CrowdHuman and CityPersons datasets show that the proposed algorithm serves as an efficient solution to handle the dense pedestrian detection problem for the case of single-stage detectors. Our method can also be flexibly applied to two-stage detectors, achieving notable performance enhancement.
\end{abstract}

\section{Introduction}
Pedestrian detection in a crowd, as a specific branch of object detection, has been widely studied in recent years \cite{lu2019semantic, chi2020pedhunter, chi2020relational, zhang2019double,zhang2018occlusion, liu2019adaptive, chu2020detection, wang2017repulsion} due to massive applications.
However, heavy occlusion and high overlap among human instances make extracting instance-wise object bounding boxes a challenging task for pedestrian detectors.

A lot of deep learning based object detectors have been proposed in the past few years in this field, and they are typically categorized into two-stage detectors and single-stage detectors. Single-stage methods show fantastic efficiency and performance for general-purpose object detection. These detectors, in brief, operates as follows: an image $\bf f$ is first passed through a fully convolutional network to predict 
a set of dense proposals $\bf z$. A post-processing step, which typically includes a non-maximum suppression (NMS) and a score threshold, is then applied to predict the final detection results $\bf x$.

\begin{figure}[t!]
  \centering
  \begin{subfigure}{.12\textwidth}

    \centering
    \begin{tikzpicture}[scale=0.6]

      \filldraw[rounded corners, fill=black!2] (-1, 1) rectangle (1, -5);
      \path (0, 0) node[thick, draw, shape=circle, minimum size=1.5em] (x1) {$\bf f$}
      (0, -2) node[thick, draw, shape=circle, minimum size=1.5em, fill=black!20] (z1) {$\bf z$}
      (0, -4) node[thick, draw, shape=circle, minimum size=1.5em] (x2) {$\bf x$};
      \draw[thick, ->, >=stealth] (x1) -- (z1) node[midway, anchor=west] {$p_\theta$};
      \draw[thick, <-, >=stealth] (z1) -- (x2) node[midway, anchor=west] {gt};

    \end{tikzpicture}
    \caption{Offline.}
    \label{fig:offline}
  \end{subfigure}
  \begin{subfigure}{.12\textwidth}

    \centering
    \begin{tikzpicture}[scale=0.6]

      \filldraw[rounded corners, fill=black!2] (-1, 1) rectangle (1, -5);
      \path (0, 0) node[thick, draw, shape=circle, minimum size=1.5em] (x1) {$\bf f$}
      (0, -2) node[thick, draw, shape=circle, minimum size=1.5em, fill=black!20] (z1) {$\bf z$}
      (0, -4) node[thick, draw, shape=circle, minimum size=1.5em] (x2) {$\bf x$};
      \draw[thick, ->, >=stealth] (x1) -- (z1) node[midway, anchor=west] {$p_\theta$};
      \draw[thick, ->, >=stealth] (z1) .. controls (-0.25, -3) .. (x2) node[midway, anchor=east] {};
      \draw[thick, <-, >=stealth] (z1) -- (x2) node[midway, anchor=west] {gt};

    \end{tikzpicture}
    \caption{Online.}
    \label{fig:online}
  \end{subfigure}
  \begin{subfigure}{.12\textwidth}

    \centering
    \begin{tikzpicture}[scale=0.6]

      \filldraw[rounded corners, fill=black!2] (-1, 1) rectangle (1, -5);
      \path (0, 0) node[thick, draw, shape=circle, minimum size=1.5em] (x1) {$\bf f$}
      (0, -2) node[thick, draw, shape=circle, minimum size=1.5em] (z1) {$\bf z$}
      (0, -4) node[thick, draw, shape=circle, minimum size=1.5em, fill=black!20] (x2) {$\bf x$};
      \draw[thick, ->, >=stealth, dashed] (x1) -- (z1) node[midway, anchor=west] {$q_\phi$};
      \draw[thick, ->, >=stealth] (z1) -- (x2) node[midway, anchor=west] {$\widetilde{p}$};

    \end{tikzpicture}
    \caption{Ours.}
    \label{fig:ours}
  \end{subfigure}
  \begin{subfigure}{.5\textwidth}
    \centering

    \begin{tikzpicture}[scale=0.57]
      \draw node [anchor=mid] (image) at (-1,0) {\includegraphics[width=0.4in]{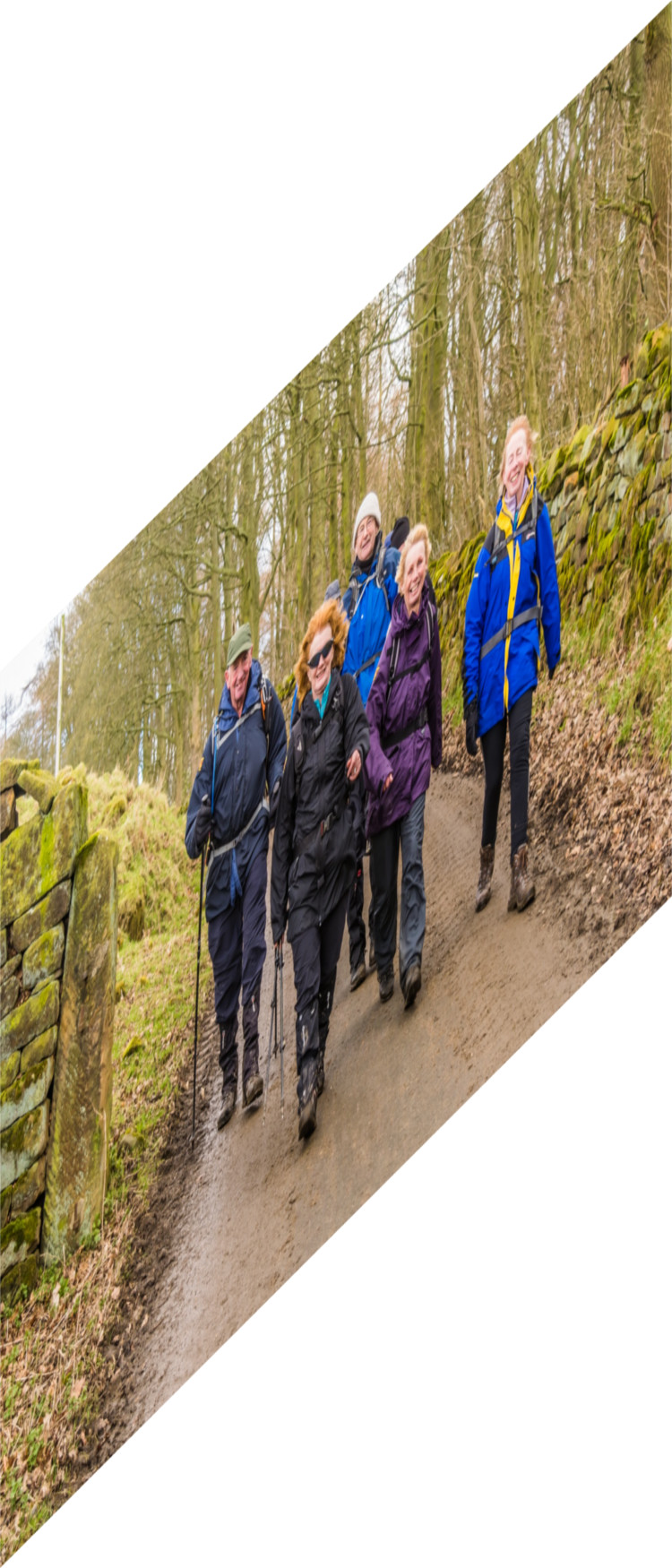}};
      \path (-1, -0.5) node {$\mathbf{f}$} +(4, 0)node  {$\mathbf{z}$} +(8,0) node {$\mathbf{x}$};
      \path (-1, 5) node {\small Image} +(4, 0)node  {\small Dense Proposals} +(8,0) node {\small Final Boxes};

      \draw[red!37] (2.274,0.028) \rect{0.303}{0.968};
      \draw[red!8] (2.417,0.211) \rect{0.398}{0.899};
      \draw[red!7] (2.599,0.345) \rect{0.259}{0.904};
      \draw[red!17] (2.68,0.452) \rect{0.339}{0.945};
      \draw[red!53] (2.833,0.584) \rect{0.332}{0.925};
      \draw[red!19] (2.966,0.831) \rect{0.362}{0.969};
      \draw[red!18] (3.087,1.043) \rect{0.322}{0.922};
      \draw[red!42] (3.282,1.047) \rect{0.244}{0.822};
      \draw[red!38] (3.49,1.315) \rect{0.341}{0.906};
      \draw[red!46] (3.511,1.352) \rect{0.296}{0.912};
      \draw[red!37] (2.188,0.465) \rect{0.266}{1.005};
      \draw[red!32] (2.443,0.646) \rect{0.34}{0.808};
      \draw[red!33] (2.571,0.6) \rect{0.303}{0.904};
      \draw[red!18] (2.776,0.786) \rect{0.345}{0.902};
      \draw[red!13] (2.915,0.952) \rect{0.267}{0.865};
      \draw[red!24] (3.057,1.031) \rect{0.336}{1.025};
      \draw[red!38] (3.133,1.28) \rect{0.274}{0.855};
      \draw[red!54] (3.32,1.365) \rect{0.245}{0.977};
      \draw[red!8] (3.38,1.446) \rect{0.231}{0.907};
      \draw[red!37] (3.602,1.67) \rect{0.329}{0.88};
      \draw[red!14] (2.206,0.67) \rect{0.381}{0.919};
      \draw[red!36] (2.382,0.89) \rect{0.299}{0.925};
      \draw[red!28] (2.612,0.97) \rect{0.344}{0.88};
      \draw[red!58] (2.706,1.071) \rect{0.284}{0.904};
      \draw[red!60] (2.814,1.196) \rect{0.225}{0.876};
      \draw[red!26] (2.952,1.338) \rect{0.336}{0.922};
      \draw[red!29] (3.153,1.616) \rect{0.355}{0.883};
      \draw[red!16] (3.299,1.737) \rect{0.279}{0.886};
      \draw[red!51] (3.492,1.867) \rect{0.191}{0.939};
      \draw[red!26] (3.57,1.926) \rect{0.246}{0.899};
      \draw[red!36] (2.202,0.964) \rect{0.304}{0.887};
      \draw[red!36] (2.393,1.121) \rect{0.291}{0.912};
      \draw[red!20] (2.593,1.251) \rect{0.43}{0.978};
      \draw[red!50] (2.686,1.346) \rect{0.263}{1.001};
      \draw[red!25] (2.844,1.594) \rect{0.329}{0.841};
      \draw[red!67] (2.966,1.77) \rect{0.302}{0.886};
      \draw[red!52] (3.142,1.9) \rect{0.296}{0.944};
      \draw[red!38] (3.337,2.023) \rect{0.409}{0.91};
      \draw[red!25] (3.375,2.137) \rect{0.314}{0.859};
      \draw[red!46] (3.496,2.29) \rect{0.337}{0.836};
      \draw[red!34] (2.236,1.233) \rect{0.301}{0.921};
      \draw[red!14] (2.407,1.403) \rect{0.254}{0.833};
      \draw[red!36] (2.611,1.477) \rect{0.314}{0.926};
      \draw[red!60] (2.69,1.674) \rect{0.221}{0.894};
      \draw[red!22] (2.822,1.845) \rect{0.292}{0.969};
      \draw[red!39] (3.076,2.018) \rect{0.234}{0.985};
      \draw[red!65] (3.137,2.135) \rect{0.293}{0.964};
      \draw[red!18] (3.285,2.403) \rect{0.317}{0.821};
      \draw[red!43] (3.453,2.446) \rect{0.381}{0.845};
      \draw[red!30] (3.582,2.572) \rect{0.279}{0.879};
      \draw[red!33] (2.245,1.573) \rect{0.269}{0.807};
      \draw[red!7] (2.368,1.773) \rect{0.291}{0.894};
      \draw[red!28] (2.509,1.835) \rect{0.279}{1.015};
      \draw[red!13] (2.755,1.858) \rect{0.397}{0.918};
      \draw[red!24] (2.98,2.184) \rect{0.294}{0.892};
      \draw[red!31] (3.061,2.365) \rect{0.315}{1.038};
      \draw[red!60] (3.145,2.532) \rect{0.293}{1.025};
      \draw[red!52] (3.322,2.617) \rect{0.314}{0.941};
      \draw[red!36] (3.423,2.781) \rect{0.283}{0.875};
      \draw[red!44] (3.565,2.906) \rect{0.322}{0.893};

      \draw[red!80] (3, 0) ++(-0.35,1) \rect{0.36}{1.25};
      \draw[red!90] (3, 0) ++(-0.1,1) \rect{0.32}{1.4};
      \draw[red!80] (3, 0) ++(0,1.4) \rect{0.303}{1.3};
      \draw[red!99] (3, 0) ++(0.1,1.35) \rect{0.32}{1.25};
      \draw[red!80] (3, 0) ++(0.35,1.55) \rect{0.4}{1.3};
      \draw[red!80] (3.01, 0.05) ++(-0.35,1) \rect{0.36}{1.25};
      \draw[red!90] (3.01, 0.05) ++(-0.1,1) \rect{0.32}{1.4};
      \draw[red!80] (3.01, 0.05) ++(0,1.4) \rect{0.303}{1.3};
      \draw[red!99] (3.01, 0.05) ++(0.1,1.35) \rect{0.32}{1.25};
      \draw[red!80] (3.01, 0.05) ++(0.35,1.55) \rect{0.4}{1.3};
      \draw[red!80] (2.98, -0.05) ++(-0.35,1) \rect{0.36}{1.25};
      \draw[red!90] (2.98, -0.05) ++(-0.1,1) \rect{0.32}{1.4};
      \draw[red!80] (2.98, -0.05) ++(0,1.4) \rect{0.303}{1.3};
      \draw[red!99] (2.98, -0.05) ++(0.1,1.35) \rect{0.32}{1.25};
      \draw[red!80] (2.98, -0.05) ++(0.35,1.55) \rect{0.4}{1.3};

      \draw[blue] (7, 0) ++(-0.35,1) \rect{0.36}{1.25};
      \draw[blue] (7, 0) ++(-0.1,1) \rect{0.32}{1.4};
      \draw[blue] (7, 0) ++(0,1.4) \rect{0.303}{1.3};
      \draw[blue] (7, 0) ++(0.1,1.35) \rect{0.32}{1.25};
      \draw[blue] (7, 0) ++(0.35,1.55) \rect{0.4}{1.3};

      \draw[thick, black] (-1, 0) ++(-0.89,-0.09) \rect{3.56}{2.36};
      \draw[thick, black] (3, 0) ++(-0.89,-0.09) \rect{3.56}{2.36};
      \draw[thick, black] (7, 0) ++(-0.89,-0.09) \rect{3.56}{2.36};

      \filldraw[black, fill=cyan!4] (0.25, 2.8) -- (1.5, 2.3) -- (1.5, 1.3) -- (0.25, 0.8) -- (0.25, 2.8)  (0.875, 1.8) node {$q_\phi$};
      \draw[->, very thick] (4.25, 1.8) -- ++(1.25, 0) node[midway, above] {$\widetilde{p}$};

    \end{tikzpicture}
    \caption{The three variables in single-stage object detection.}
    \label{fig:threecmp}
  \end{subfigure}
  \caption{\textbf{Graph illustration of relation modeling in single-stage object detection.} Object detectors optimized by: (a) {offline} ground truth assignment \cite{liu2016ssd,redmon2016you,lin2020focal,tian2019fcos}; (b) online ground truth assignment \cite{zhang2019freeanchor,liu2019hambox}; (c) Variational Pedestrian detector. Dashed line denote variational models; the variable of interest is marked gray. }
  \label{fig:concept}
  \vspace{-2ex}
\end{figure}
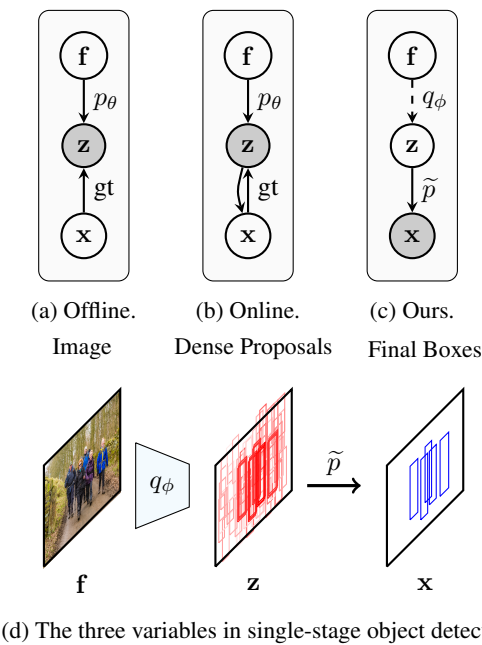

Single-stage object detectors are usually trained over an IoU related loss between pre-defined dense boxes (namely \emph{anchors}) $\bf z$ and ground truth boxes. We define this learning \textbf{offline} as shown in \autoref{fig:offline}; the method first assigns the ground truth to anchors and then adjusts the anchor boxes through regression. This may be ambiguous in crowded scenes whereby a single anchor usually highly overlaps with multiple object instances \cite{chabot2019lapnet}. Thus, this could yield a sub-optimal solution and greatly hinders the performance. To handle this issue, a series of methods \cite{zhang2019freeanchor, liu2019hambox} have been proposed to adjust the object proposal before, or even simultaneously with the assignment procedure. We call this kind of learning as \textbf{online}, as illustrated in \autoref{fig:online}. However, online pipelines still conform to a certain handcrafted matching rules and result in less sufficient exploration of matching space between proposals and groundtruth.


Different from online and offline methods where the dense proposal $\bf z$ is considered as part of the optimization target, we here formulate the dense proposal $\bf z$ as an auxiliary latent variable which relates to the final detection $\bf x$ as our target (as illustrated in \autoref{fig:ours} and \autoref{fig:threecmp}). To be specific, we introduce a random learning-to-match approach via variational inference, which predicts the distribution of dense proposals $\bf z$ instead of deterministic value. Such variational pedestrian detector can learn to adaptively adjust the exploration strategy in matching space by itself, thus can handle heavy occlusion for pedestrian detection when only training with full body information. Another important property is the plug-and-play nature which makes our method cater to both anchor-based and anchor-free detection pipelines.

The major contributions in this paper are three-fold:
\begin{itemize}[itemsep= 0pt,topsep = 0pt,partopsep=0pt]
  \item We propose a a brand new perspective of formulating single-stage detectors as a variational inference problem and intend to motivate further extensions of detection schemes along this direction.
  \item We introduce a detection-customized Auto-Encoding Variational Bayes (AEVB) algorithm motivated by \cite{kingma2014auto} to optimize the general object detectors for pedestrian detection.
  \item Our experiments on CrowdHuman \cite{shao2018crowdhuman} and CityPersons \cite{zhang2017citypersons, cordts2016the} datasets demonstrate the effectiveness of the proposed approach for single-stage and two-stage detectors.
\end{itemize}

\section{Related works}

\subsection{Object Detection}
Object detection aims to find a set of boxes $\bf x$ indicating the objects in the image $\bf f$. Recent object detection works are mainly either single-stage or two-stage methods.

\textbf{Single-stage object detection methods} adopt a fully convolutional network (FCN) to map dense proposals $\bf z$ to final sparse detection boxes $\bf x$.
It typically connects FCN with two heads: a classification head to predict the existence of a local object, and a regression head to refine the bounding box of the object.
Single-stage detectors can be further categorized into \textit{anchor-based} methods \cite{liu2016ssd, redmon2018yolov3, lin2020focal} and \textit{anchor-free} methods \cite{liu2019high,tian2019fcos,law2020cornernet}. Anchor-based methods defined dense proposals with default boxes, while anchor-free methods use pixels as an alternative to anchors, and predicts the object class and bounding box for each pixel \cite{liu2019high,tian2019fcos} or pair of pixels \cite{law2020cornernet}.

\textbf{Two-stage object detection methods}, represented by Faster R-CNN \cite{ren2017faster} and its variants, typically feature a region proposal network (RPN) and a region-based convolutional neural network (R-CNN)~\cite{girshick2014rich, girshick2015fast}. The first stage generates object proposals by the RPN, and the second stage proceeds to refine and predict object category for each proposal. Two-stage methods almost dominate pedestrian detection research due to their good performance~\cite{zhang2016is, hosang2017learning, hu2018relation, liu2019adaptive, chu2020detection}.

\textbf{Pedestrian detection} has greatly advanced together with the progress of general object detection. However, pedestrian detection has its distinct challenges, particularly the issue of occlusion in crowded scenarios.
Specific techniques were proposed in the past to improve detection performance, such as detecting body by parts \cite{lu2019semantic, chi2020pedhunter, chi2020relational, zhang2019double,zhang2018occlusion}, improving non-maximum suppression \cite{chi2020relational, zhang2019double, liu2019adaptive, chu2020detection, wang2017repulsion}, and redesigning the anchors \cite{zhang2019double,chu2020detection, zhang2017citypersons}.
Most of these are two-stage methods, while single-stage approaches \cite{liu2019high, liu2018learning} usually require complex structures to catch up on the performance of two-stage methods.

\subsection{Online Anchor Matching}
The aforementioned detection methods perform the assignment of ground truth to anchors before adjusting the object boxes. This \textbf{offline} procedure is likely to introduce ambiguity, especially for highly occluded crowded scenarios.
On the contrary, \textbf{online} anchor matching adjusts/predicts object boxes first before assigning ground truth boxes to dense proposals. One such recent work is HAMBox \cite{liu2019hambox}, which implements a high-quality online anchor mining strategy to assign positive and negative samples. FreeAnchor \cite{zhang2019freeanchor} formulate the assignment and proposal regression in one framework as a maximum likelihood estimation procedure.
Concretely, both online and offline methods take dense proposal $\bf z$ (instead of the final target $\bf x$) as the optimization target in the loss function, which is sub-optimal for the object detection task. Instead, we hypothesize that the target $\bf x$ can be directly optimized by reformulating the entire detection task as a variational inference problem, treating the dense proposal $\bf z$ as a latent auxiliary variable.

\subsection{Auto-Encoding Variational Bayes}
The Auto-Encoding Variational Bayes \cite{kingma2014auto} (AEVB) algorithm enables efficient training of probabilistic models with latent variables using a reparameterization trick. It can be used to approximate posterior inference of the latent variable $\bf z$ given an observed value $\bf x$, which is useful for coding or data representation tasks \cite{kingma2014auto}. Our model is closely related to the variational auto-encoder (VAE) in the sense that we learn the dense proposal as a latent variable. The dense proposal model encodes the image to the dense proposal space where the decoder model extracts the detection boxes from the ``code''.

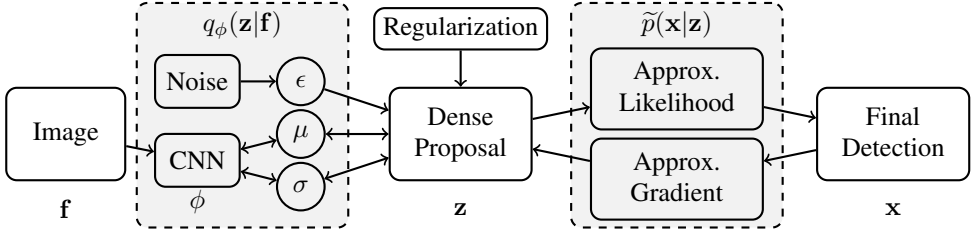
\begin{figure*}[ht]
  \centering
  \begin{tikzpicture}
    \path (2.35 , 0.6) node[thick, draw, minimum height=8.5em, minimum width=8em, shape=rectangle, rounded corners, fill=black!5, dashed] (q) {} node[anchor=north] at (q.north) {$q_\phi(\bf z|f)$};
    \path (8.125 , 0.6) node[thick, draw, minimum height=8.5em, minimum width=8em, shape=rectangle, rounded corners, fill=black!5, dashed] (q) {} node[anchor=north] at (q.north) {$\widetilde{p}(\bf x|z)$};
    \path (0 , -0.65) node{$\bf f$} (5.25 , -0.65) node{$\bf z$} (11 , -0.65) node{$\bf x$};

    \path (0,0.35) node[thick, draw, minimum height=3.5em, minimum width=4.5em, shape=rectangle, rounded corners] (f) {Image}

    (1.75,1.05) node[thick, draw, minimum height=2em, minimum width=3.2em, shape=rectangle, rounded corners] (n) {Noise}
    (1.75,0) node[thick, draw, minimum height=2em, minimum width=3.2em, shape=rectangle, rounded corners] (cnn) {CNN}
    node[anchor=north] at ([yshift=3pt]cnn.south) {$\phi$}

    (3.125, 0.35) node[thick, draw, shape=circle, minimum size=1.8em] (m) {$\mu$}
    (3.125,-0.35) node[thick, draw, shape=circle, minimum size=1.8em] (s) {$\sigma$}
    (3.125,1.05) node[thick, draw, shape=circle, minimum size=1.8em] (e) {$\epsilon$}
    (5.25,0.35) node[thick, draw, minimum height=3.5em, minimum width=4.5em, shape=rectangle, rounded corners] (z) {\begin{tabular}{c}Dense \\ Proposal\end{tabular}}
    (5.25,1.75) node[thick, draw, shape=rectangle, rounded corners] (r) {Regularization}

    (8.125, 0.95) node[thick, draw, minimum width=6.5em, shape=rectangle, rounded corners] (zx) {\begin{tabular}{c}Approx. \\ Likelihood\end{tabular}}
    (8.125,-0.25) node[thick, draw, minimum width=6.5em, shape=rectangle, rounded corners] (xz) {\begin{tabular}{c}Approx. \\ Gradient\end{tabular}}
    (11,0.35) node[thick, draw, minimum height=3.5em, minimum width=4.5em, shape=rectangle, rounded corners] (x) {\begin{tabular}{c}Final \\ Detection\end{tabular}};


    \draw[thick, ->] (f) -- (cnn);
    \draw[thick, ->] (n) -- (e);
    \draw[thick, ->] (e) -- (z);
    \draw[thick, ->] (r) -- (z);
    \draw[thick, <->] (cnn) -- (m);
    \draw[thick, <->] (cnn) -- (s);
    \draw[thick, <->] (m) -- (z);
    \draw[thick, <->] (s) -- (z);
    \draw[thick, ->] (zx) -- (x);
    \draw[thick, <-] (xz) -- (x);
    \draw[thick, ->] (z) -- (zx);
    \draw[thick, <-] (z) -- (xz);

  \end{tikzpicture}
  \caption{\textbf{Illustration of auto-encoding variational bayes (AEVB) algorithm for pedestrian detection.} From the perspective of variational auto-encoder, the final detection $\bf x$ is represented by the dense proposal $\bf z$. The reparameterization trick represents the latent random variable $\mathbf{z}$ by an independent auxiliary random variable $\epsilon$ and an invertible function $g_\phi: \epsilon \mapsto \mathbf{z}$ parameterized by the variational parameter $\phi$.}
  \label{fig:aevb}
  \vspace{-2ex}
\end{figure*}

\section{Method}

\subsection{Problem Setup}
Given an object detection dataset $\left\{\left(\mathbf{f}^{(i)}, \mathbf{x}^{(i)}\right)\right\}_{i=1}^N$ which consists of $N$ i.i.d. pairs of image $\bf f$ and a set of object bounding boxes $\bf x$, we want to predict object bounding boxes $\bf x$ for a new input image $\bf f$. To deal with an unknown number of instances in the image, we introduce the auxiliary variable $\bf z$. As illustrated in \autoref{fig:ours} and \autoref{fig:threecmp}, we perform single-stage object detection through a variational model $p(\bf x | f)$ that integrate seamlessly two probabilistic modules: a variational dense proposal generation module $q_\phi(\bf z|f)$ parameterized by a convolutional neural network, and the final detection extraction module $p(\bf x|z)$.

The formulation, succinctly, is as follows: \textit{First}, $q_\phi(\bf z|f)$ encodes $\bf f$ to output dense proposal $\bf z$.
\textit{Then}, $p(\bf x|z)$ predicts the final detection $\bf x$. %
%
The two modules are seamlessly integrated
such that it is entirely different from the aforementioned online and offline methods, which regard the dense proposal $\bf z$ as the variable of interest.
The variational model is solved with a customized AEVB algorithm and a pseudo detection likelihood as illustrated in \autoref{fig:aevb} and described in the following sections.

\subsection{Variational Detection}
We approximate the true posterior $p(\mathbf{x}|\mathbf{f})$ by a variational distribution $q_\phi(\mathbf{z}|\mathbf{f})$ with variational parameters $\phi$. Assuming the datapoints $\bf f,x$ are independent, we use $\bf f,x$ to denote one sample in the dataset. The log likelihood for final detection $\bf x$ can be written as:
\begin{equation}
  \small
  \log p(\mathbf{x}) = \int q_\phi(\mathbf{z}|\mathbf{f})\log \frac{p(\mathbf{x}, \mathbf{z})}{q_\phi(\mathbf{z}|\mathbf{f})}\mathrm{d}\mathbf{z} + KL(q_\phi(\mathbf{z}|\mathbf{f})\|p(\mathbf{z}|\mathbf{x})),
  \label{eq:logpx}
\end{equation}
where the second term is the Kullback-Leibler (KL) divergence, which is a measure of similarity between two distributions. Since the KL divergence is non-negative, the first term is called the evidence lower bound (ELBO) which can be rewritten as:
\begin{equation}
  \small
  \mathcal{L(\phi; \mathbf{f,x})} := - \alpha \cdot KL(q_\phi(\mathbf{z}|\mathbf{f})\|p(\mathbf{z})) + \mathbb{E}_{q_\phi(\mathbf{z}|\mathbf{f})}\log p(\mathbf{x}|\mathbf{z}),
  \label{eq:elbo}
\end{equation}
where scaling factor $\alpha$ is introduced to balance the scale difference between the two terms. %
An alternative explanation for the ELBO is that we want to optimize the dense proposal generation model $q_\phi(\mathbf{z}|\mathbf{f})$ such that its fitted distribution is close to the true posterior $p(\mathbf{z}|\mathbf{x})$, \emph{i.e.}, minimizing the KL term in \autoref{eq:logpx}. Since the sum of that KL term and the ELBO does not depend on the variational parameter $\phi$, maximizing the ELBO is equivalent to minimizing the KL term in \autoref{eq:logpx}.

\textbf{Analysis of ELBO terms.} 
In \autoref{eq:elbo}, the first term is the KL divergence between the variational distribution and its prior. This term can be seen as a regularization term that constrains the variational distribution of $\mathbf{z}$ to its prior. It is tractable for univariate normal distribution, and the gradient can be evaluated analytically. In practice, we impose a scale factor $\alpha$ to balance the scale differences between both ELBO terms, where the chosen value for $\alpha$ is usually small since the gradient from KL divergence is accumulated on each pixel but the relaxed detection likelihood $\widetilde{p}(\bf x|z)$ is averaged over the image.

The second term of the ELBO is often referred to as the data term:
\vspace{-0.5em}
\begin{equation}
  \small
  D(\phi; \mathbf{f,x}) := \mathbb{E}_{q_\phi(\mathbf{z}|\mathbf{f})}\log p(\mathbf{x}|\mathbf{z}),
  \label{eq:data}
\end{equation}
where $\log p(\bf x|z)$ is the detection likelihood given the known dense proposal, which can be relaxed to a tractable pseudo
detection likelihood and reflects the expected detection quality. Note that optimizing the model with plain maximum likelihood method is equivalent to approximating the data term over the expected dense proposal. Taking the expectation of the detection likelihood encourages the object detector to explore different matches and converge to a better solution.

\subsection{Optimization Algorithm}
Taking a leaf from the Auto-Encoding Variational Bayes (AEVB) \cite{kingma2014auto} algorithm, we optimize the variational single-stage object detection model using a detection-customized AEVB algorithm described in {\color{red} Algorithm} \ref{alg:req}.

The target is to find stochastic gradient for the data term (\autoref{eq:data}), \emph{i.e.} $\nabla_\phi D(\phi; \mathbf{f,x})$, to optimize the dense proposal generation model. We adopt the reparameterization gradient estimator in \cite{kingma2014auto, salimans2013fixed, ruiz2016the}. Hence, the reparameterization gradient for the data term can be rewritten as:
\begin{equation}
\small
\begin{split}
  \nabla_\phi\mathbb{E}_{q_\phi(\mathbf{z}|\mathbf{f})}&\log p(\mathbf{x}|\mathbf{z}) = \\ &\mathbb{E}_{q_\epsilon(\epsilon)}\left[\nabla_z\log p(\mathbf{x}|\mathbf{z})\rvert_{\mathbf{z}=g_\phi(\mathbf{\epsilon})}\nabla_\phi g_\phi(\mathbf{\epsilon})\right],
  \label{eq:reparametrization}
\end{split}
\end{equation}
While the data term of~\autoref{eq:reparametrization} needs to be optimized through stochastic gradient, it is non-trivial to find the probability density for a sparse set $\bf x$. Therefore, we relax the sparse detection likelihood $p(\mathbf{x}|\mathbf{z})$ to a tractable pseudo detection likelihood $\widetilde{p}(\mathbf{x}|\mathbf{z})$. The relaxed data term $\nabla_\phi\mathbb{E}_{q_\phi(\mathbf{z}|\mathbf{f})}\log \widetilde{p}(\mathbf{x}|\mathbf{z})$ will be elaborated in \autoref{plhd}.

Different from the REINFORCE gradient estimator \cite{glynn1990likelihood, williams1992simple} which can work on a non-differentiable detection likelihood (also named score function), the reparameterization gradient estimator requires the pseudo detection likelihood $\widetilde{p}(\mathbf{x}|\mathbf{z})$ to be differentiable almost everywhere.
The stochastic gradient is given by Monte Carlo estimate of the expected pseudo detection likelihood. In implementation, we apply auto-differentiation on
\begin{equation}
  \small
  \widetilde{D}(\phi; \epsilon) = \log \widetilde{p}(\mathbf{x}|\mathbf{z}) \qquad\text{where} \ \mathbf{z}=g_\phi(\mathbf{\epsilon}) \ \text{and}\  \epsilon \sim p_\epsilon(\epsilon).
  \label{eq:sdata}
\end{equation}

Similar to the widely adopted stochastic gradient descent method, AEVB computes the stochastic gradient by drawing random samples from the dataset in each iteration. However, there are two major differences: (1) the AEVB introduces additional randomness from the auxiliary variable drawn from the noise distribution; (2) the AEVB regularizes the dense proposal distribution to its prior. We analyze these two differences below separately.

\textbf{Introducing additional randomness}.
The necessity of sampling the auxiliary variable comes from the fact that our dense proposal follows a variational distribution $q_\phi(\bf z|f)$. The sampling process is essential to train all the variational parameters, \emph{e.g.}, the standard division $\sigma$ in the univariate normal distribution.
For the object detection task, the variable $\bf z$ corresponds to dense proposals on the image. Sampling $\bf z$ from a distribution can be regarded as jittering the dense proposals. An insight connected to the REINFORCE estimator is that the single-stage object detector is performing random exploration to find a better match between dense proposals and ground-truth objects by jittering the proposal boxes.
Experimentally, the standard deviation of the univariate normal distribution will gradually reduce as the model converges, which means that the random exploring space will gradually reduce and converge to the optimal match eventually.

\textbf{Regularizing dense proposals}.
Compared to the maximum likelihood estimation (the second term in \autoref{eq:elbo}), an additional regularization term $KL(q_\phi(\mathbf{z}|\mathbf{f})\|p(\mathbf{z}))$ is introduced. The behavior of the regularization depends on the choice of the prior of dense proposals. For instance, we choose the prior $p(\mathbf{z})$ to be a standard normal distribution in our implementation, which means that the regularization term will restrict the dense proposals to their anchors. Since CNNs usually perform better at extracting local features, such restriction enhances the detection by encouraging detections from the center of objects. A similar idea is applied in FCOS \cite{tian2019fcos}.
Another aspect of regularization is to restrict the variance and prevent the variational distribution from deteriorating to a point estimate so as to enhance the exploration during the training stage.
  {
    \begin{algorithm}[tp!]
    \small
      \caption{Customized Auto-Encoding Variational Bayes (AEVB) algorithm for object detection}
      \label{alg:req}
      \begin{algorithmic}[1]
        \Require
        Object detection dataset $\left\{\left(\mathbf{f}^{(i)}, \mathbf{x}^{(i)}\right)\right\}_{i=1}^N$; pseudo detection likelihood $\widetilde{p}(\mathbf{x}|\mathbf{z})$; dense proposal generation CNN model $q_\phi(\mathbf{z}|\mathbf{f})$; scaling factor $\alpha$.
        \Ensure
        Model parameter $\phi$;
        \State Initialize model parameter $\phi$;
        \Repeat
        \State \parbox[t]{\dimexpr\linewidth-\algorithmicindent}{%
        Sample auxiliary variable $\epsilon$ from distribution $p_\epsilon(\epsilon)$ and minibatch $\bf f, x$ from dataset;}
        \State \parbox[t]{\dimexpr\linewidth-\algorithmicindent}{Evaluate the pseudo detection likelihood $\widetilde{p}(\mathbf{x}|\mathbf{z})$ with $\mathbf{z} = g_\phi(\epsilon)$;}
        \State \parbox[t]{\dimexpr\linewidth-\algorithmicindent}{Compute regularization gradient \hspace{8em} $\mathbf{g}_1\leftarrow\nabla_\phi KL(q_\phi(\mathbf{z}|\mathbf{f})\|p(\mathbf{z}))$;}
        \State \parbox[t]{\dimexpr\linewidth-\algorithmicindent}{Compute the stochastic gradient $\mathbf{g}_2\leftarrow\nabla_\phi \widetilde{D}(\phi; \epsilon)$ by  \autoref{eq:sdata};}
        \State \parbox[t]{\dimexpr\linewidth-\algorithmicindent}{Update model parameter $\phi$ using gradient $\alpha\cdot \mathbf{g}_1 +\mathbf{g}_2$ and optimizer (e.g. SGD)}
        \Until{convergence of model parameter $\phi$}\\
        \Return Model parameter $\phi$;.
      \end{algorithmic}
    \end{algorithm}
    \setlength{\textfloatsep}{0ex}
    \setlength{\floatsep}{0ex}
  }

\subsection{Pseudo Detection Likelihood} \label{plhd}
We have one undetermined term in the algorithm, \emph{i.e.}, the pseudo detection likelihood $\widetilde{p}(\mathbf{x}|\mathbf{z})$. Without loss of generality, we describe how  $\widetilde{p}(\mathbf{x}|\mathbf{z})$ can be defined based on FreeAnchor \cite{zhang2019freeanchor} in this section, since the pseudo detection likelihoods of other state-of-the-art detection methods are also derivable (one of them is
shown
in the \textbf{Appendix}).

FreeAnchor \cite{zhang2019freeanchor} formulates a customized object detection customized likelihood: for each ground truth box $x_i$, a bag of anchors $A_i$ is constructed to evaluate the likelihood of recalling $x_i$. Meanwhile, dense proposals $\bf z$ are regarded as negative samples by a soft IoU threshold.

Naturally, we define our pseudo final detection likelihood $\widetilde{p}(\bf x|z)$ where positive samples are obtained from ground-truth boxes in the dataset but negative samples should be mined from dense proposals. We analyze positive samples and negative samples below separately.

\textbf{Positive pseudo likelihood}.
We first define the match quality $M_{ij}$ between proposal $z_j$ and ground truth box $x_i$ by the product of classification score $z_j^{cls}$ and IoU score:
\begin{equation}
  \small
  M_{ij} := z_j^{cls}\cdot IoU(x_i,z_j).
\end{equation}
Then, the positive likelihood or recall can be defined as
\begin{equation}
  \small
  p(\mathbf{x}_{gt}=1|\mathbf{z}):=\prod_{x_i\in{\mathbf{x}_{gt}}} \max_{z_j\in\mathbf{z}}M_{ij} = \prod_{x_i\in{\mathbf{x}_{gt}}} \max_{z_j\in\mathbf{z}}(z_j^{cls}\cdot IoU(x_i,z_j)).
\end{equation}

\begin{figure}
  \small
  \centering
  \begin{subfigure}{0.23\textwidth}
    \centering
    \includegraphics[width=\textwidth, clip]{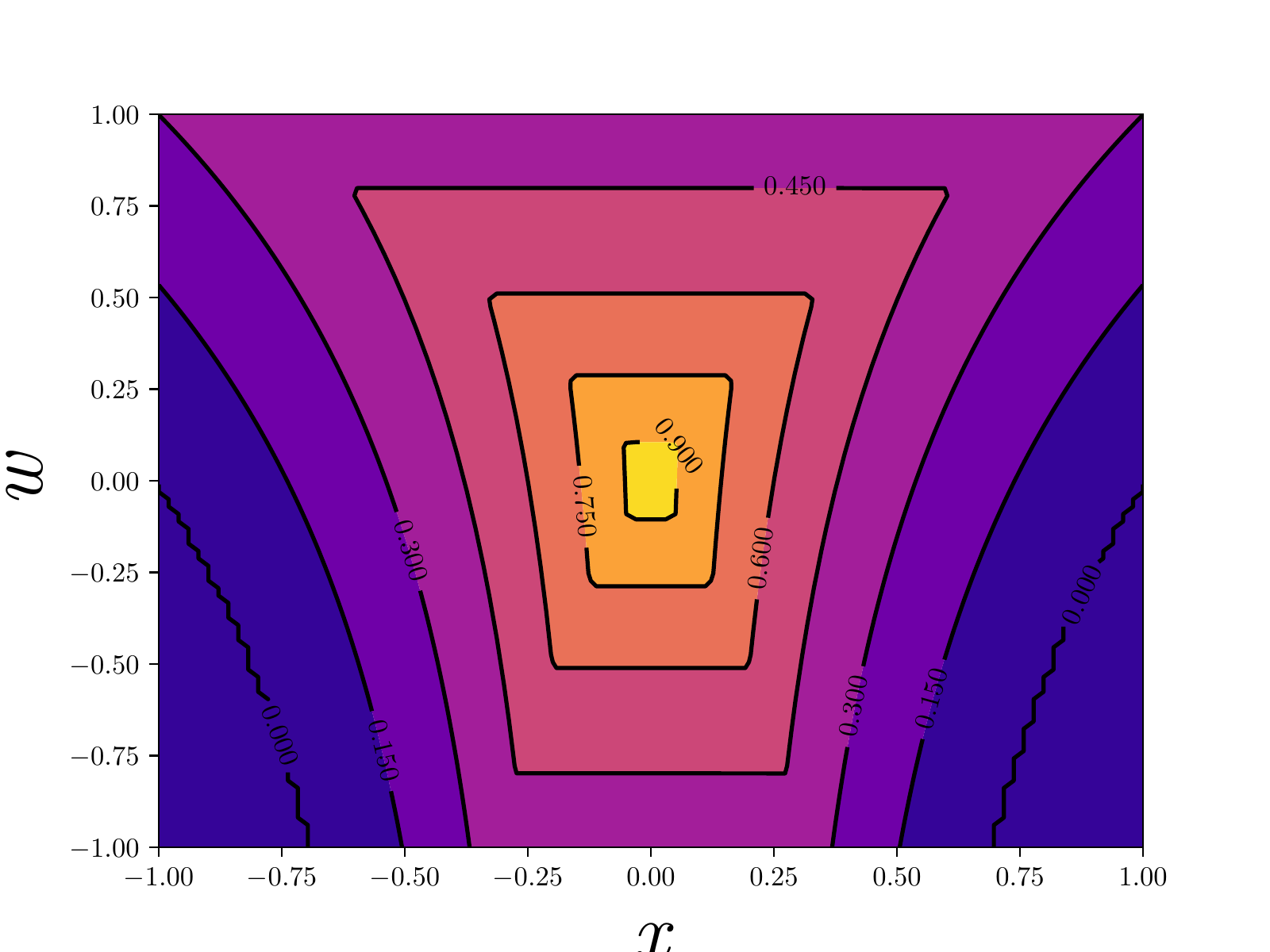}
    \caption{FA w/o AEVB}
    \label{fig:xwnos}
  \end{subfigure}
  \vspace{1pt}
  \begin{subfigure}{0.23\textwidth}
    \centering
    \includegraphics[width=\textwidth, clip]{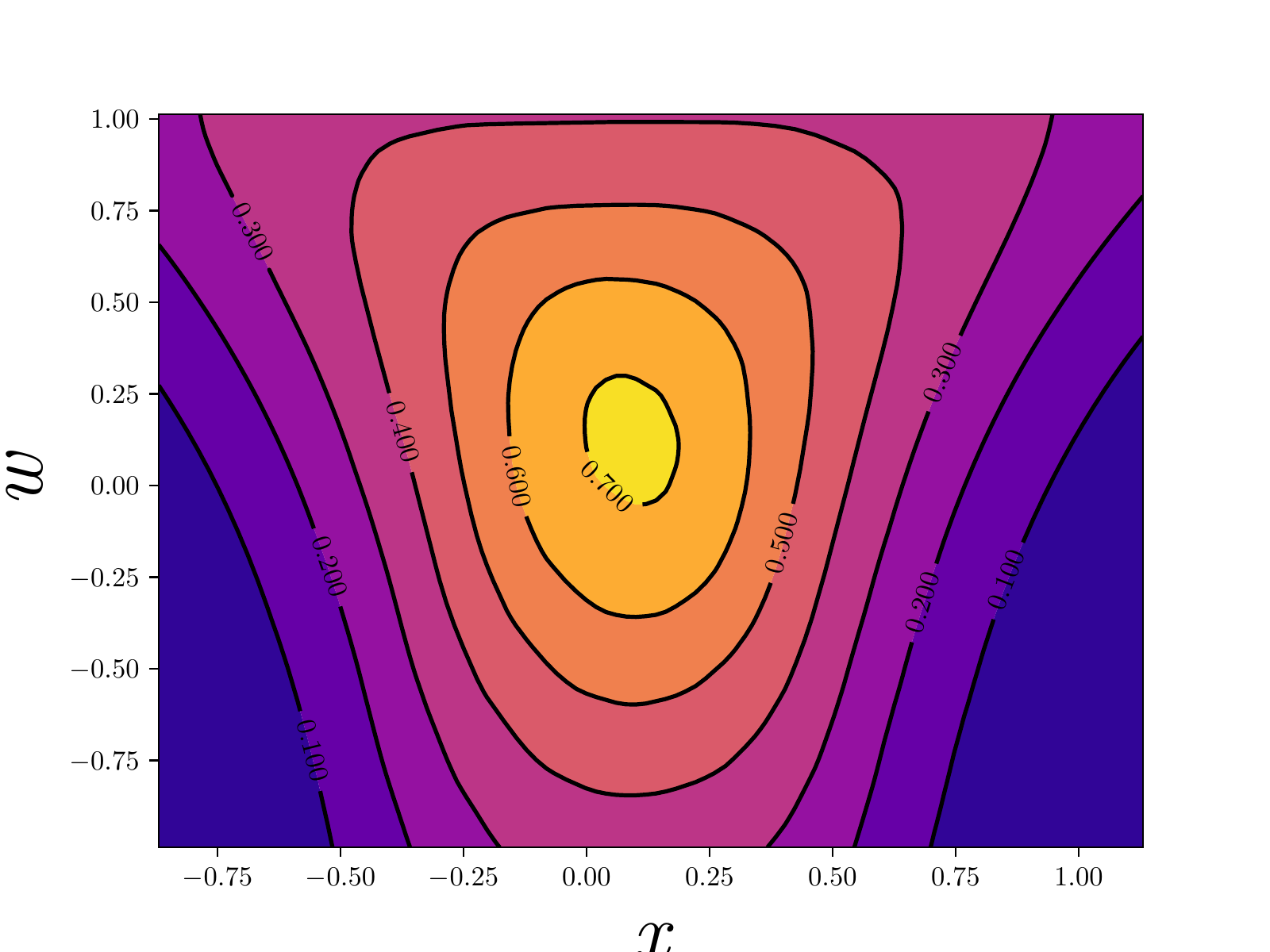}
    \caption{FA with AEVB}
    \label{fig:xwonlys}
  \end{subfigure}
  \hspace{1pt}
  \begin{subfigure}{0.23\textwidth}
    \centering
    \includegraphics[width=\textwidth, clip]{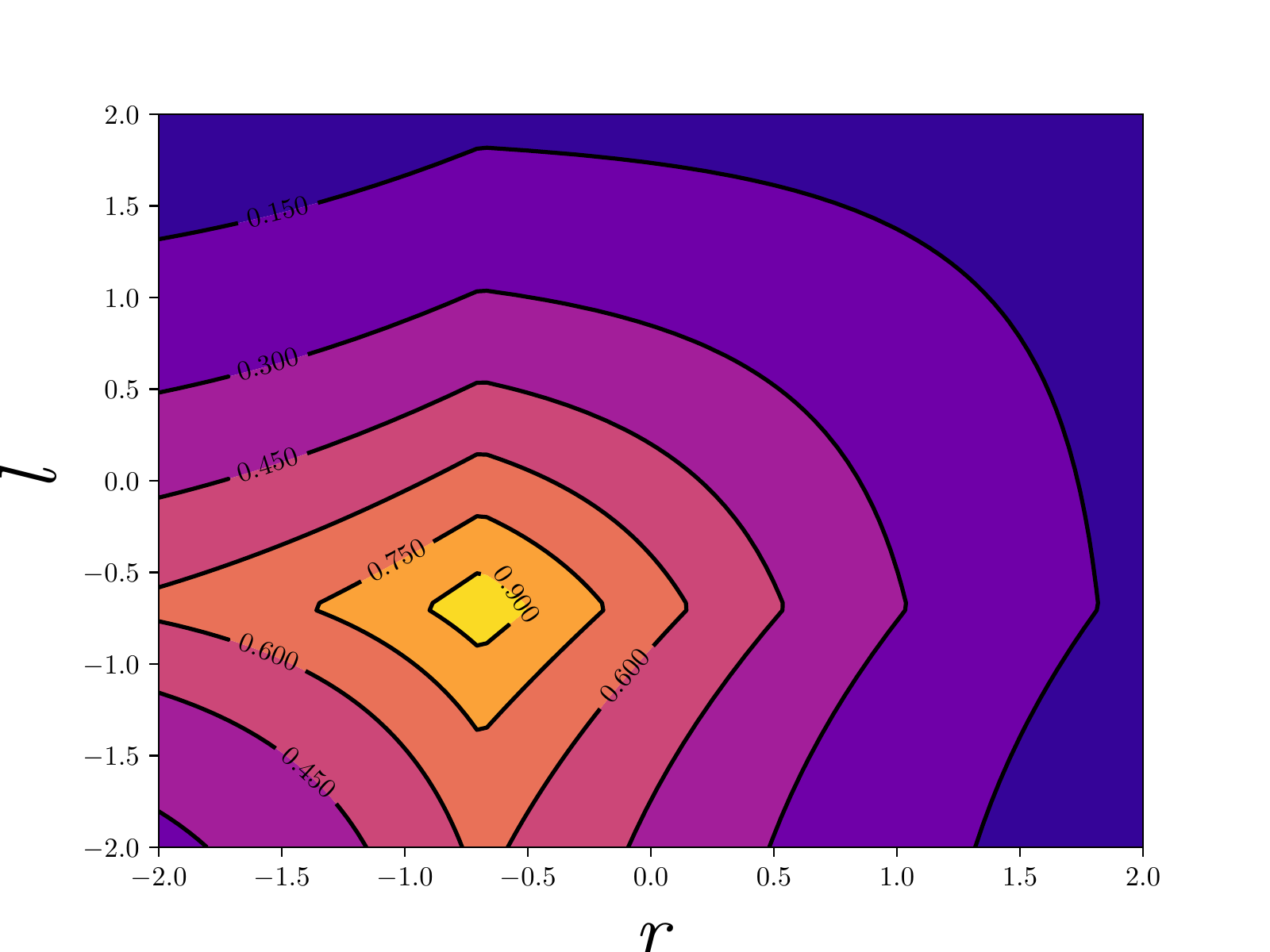}
    \caption{FCOS w/o AEVB}
    \label{fig:xxnos}
  \end{subfigure}
  \vspace{1pt}
  \begin{subfigure}{0.23\textwidth}
    \centering
    \includegraphics[width=\textwidth, clip]{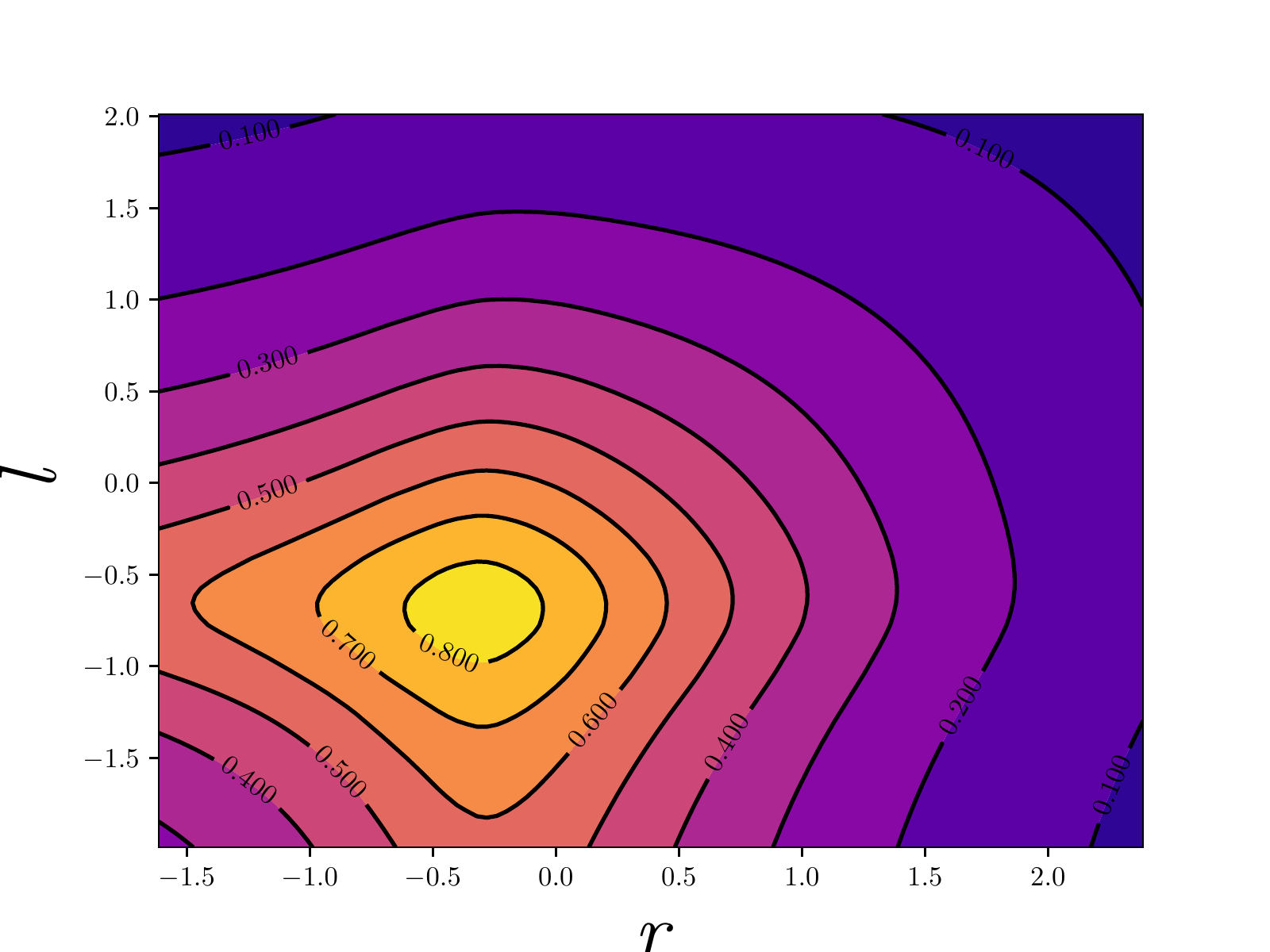}
    \caption{FCOS with AEVB}
    \label{fig:xxonlys}
  \end{subfigure}
  \caption{\textbf{The variational effect on the gradient of localization regression.} Our method can be flexibly integrated into existing single-stage detectors. For FA \cite{zhang2019freeanchor}, the regression targets are location, width and height. The IoU contour maps are plotted along the $x$ (horizontal location) and $w$ (width) axis. The ground truth box is located at origin with unit width and height. The contour maps of FCOS \cite{tian2019fcos} are similarly plotted where $r$, $l$ represent distances from the location to the right side and left side, respectively.}
  \label{fig:visualgrad}
  \vspace{-2ex}
\end{figure}

The mean-max function in \cite{zhang2019freeanchor} is proposed as a smooth relaxation to the hard-max for efficient training. The notation $z_j\in A_i$ indicates the top $n$ dense proposals $z_j$ with highest IoU to ground truth box $x_i$, and the positive pseudo likelihood is computed as
\begin{equation}
  \small
  \mathcal{P}_{i, pos} := \text{Mean-max}(x_i|\mathbf{z}) = \frac{\sum_{z_j\in A_i}\frac{M_{ij}}{1-M_{ij}}}{\sum_{z_j\in A_i}\frac{1}{1-M_{ij}}}.
  \label{eq:meanmax}
\end{equation}
Although the form of positive likelihood is similar to that in \cite{zhang2019freeanchor}, we replace the localization term with the IoU that is used in $M_{ij}$. This is due to three facts.
First, the IoU loss has shown relatively better performance, as in UnitBox \cite{yu2016unitbox}.
Second, IoU is hyperparameter-free and can be considered as a non-parametric localization likelihood.
Third, from the perspective of optimization, the IoU loss is more compatible to our proposed method. In \autoref{fig:visualgrad} we show the variational effect on gradient from localization regression. By applying variational inference, we properly smooth the gradient at some angular points, making the learning process more stable. Experiments in \autoref{exp} also suggest that when IoU loss and variational detector are used together, our method achieves better results for crowded pedestrian detection.

\textbf{Negative pseudo likelihood}
cannot be directly constructed by sampling ground truth boxes from the dataset. It is necessary to involve a probabilistic negative sample mining procedure for efficiency consideration.

We define $1-s_j:= P[z_j=0]$, which corresponds to the precision likelihood in \cite{zhang2019freeanchor}. Hence, $s_j$ represents the probability of failing to suppress the negative proposal $z_j$.
The maximum likelihood method implies binary cross-entropy loss for dense proposal $\bf z$. However, recent studies \cite{lin2020focal,chen2019towards,shrivastava2016training,zhang2017single} show that single-stage detectors optimized by cross-entropy loss suffer from extreme foreground and background imbalance which results in poor performance. To handle this problem, Focal loss \cite{lin2020focal} remains one of the state-of-the-art methods.

In this paper, we reformulate Focal loss \cite{lin2020focal} as a probabilistic online hard example mining \cite{shrivastava2016training} method to obtain negative samples for computing the negative likelihood. Recall that not all boxes $z_j$ will be counted as negative samples in the final detection $\bf x$ since (1) low-score boxes $z_j$ are usually ignored in practice; (2) ranking based metrics like average precision (AP) are more sensitive to high-score boxes.
Formally, we introduce an independent auxiliary Bernoulli random variable $K_j$ with parameter $k_j:=P[K_j=1]$. We choose to include $z_j$ as a negative sample if and only if $K_j=1$. Thus, the negative likelihood is:
\begin{equation}
  \small
  \mathcal{P}_{j, neg} := P[z_jK_j=0] = 1 - P[z_jK_j=1] = 1 - k_j s_j.
  \label{eq:fpl}
\end{equation}

By designing the probability to keep the dense proposal as $k_j:=(1 - (1 - s_j)^{s_j^\gamma})/s_j$,
the negative pseudo likelihood is reduced to the Focal Loss \cite{lin2020focal}. A detailed proof is provided in the \textbf{Appendix}.

\begin{figure*}[t]
  \small
  \centering
  \begin{subfigure}{0.23\textwidth}
    \begin{minipage}{\textwidth}
    \centerline{\includegraphics[width=\textwidth]{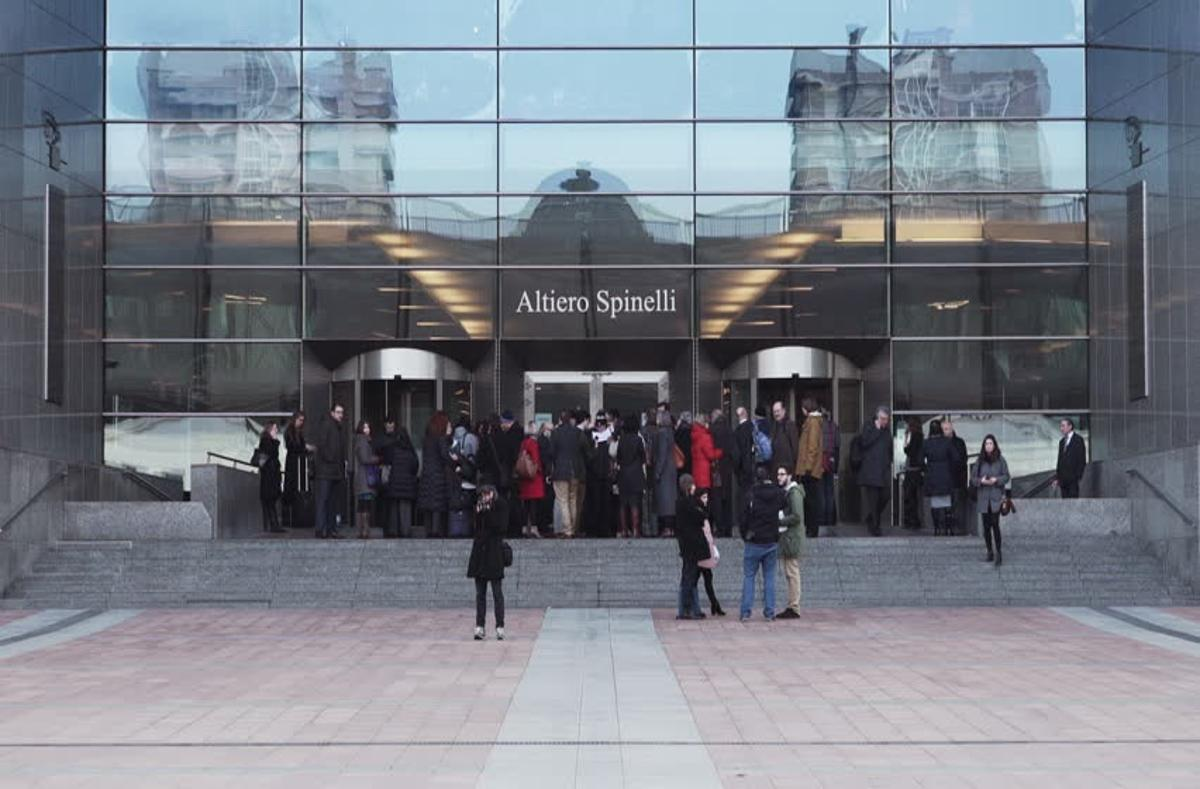}}\vspace{1pt}
    \centerline{\includegraphics[width=\textwidth]{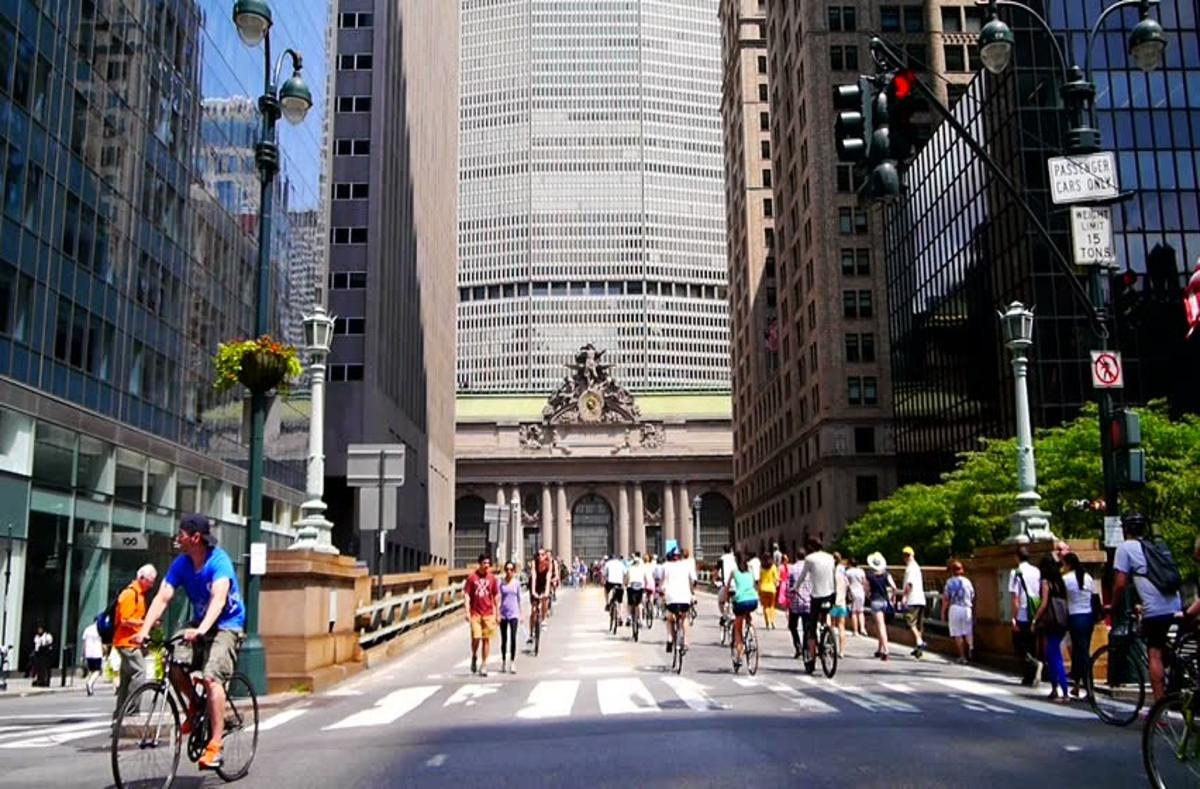}}\vspace{1pt}
    \end{minipage}
    \caption{Input image.}
    \label{fig:input}
  \end{subfigure}
  \hspace{1pt}
  \begin{subfigure}{0.23\textwidth}
    \begin{minipage}{\textwidth}
    \centerline{\includegraphics[width=\textwidth]{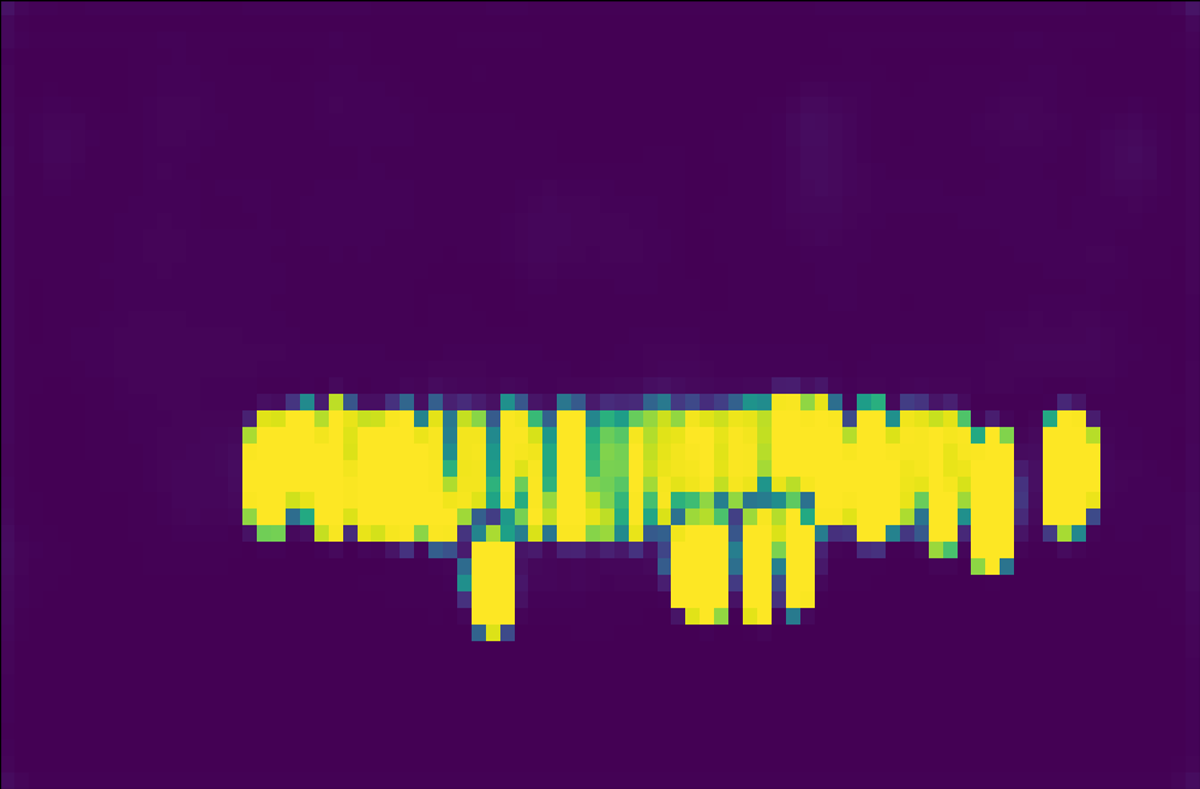}}\vspace{1pt}
    \centerline{\includegraphics[width=\textwidth]{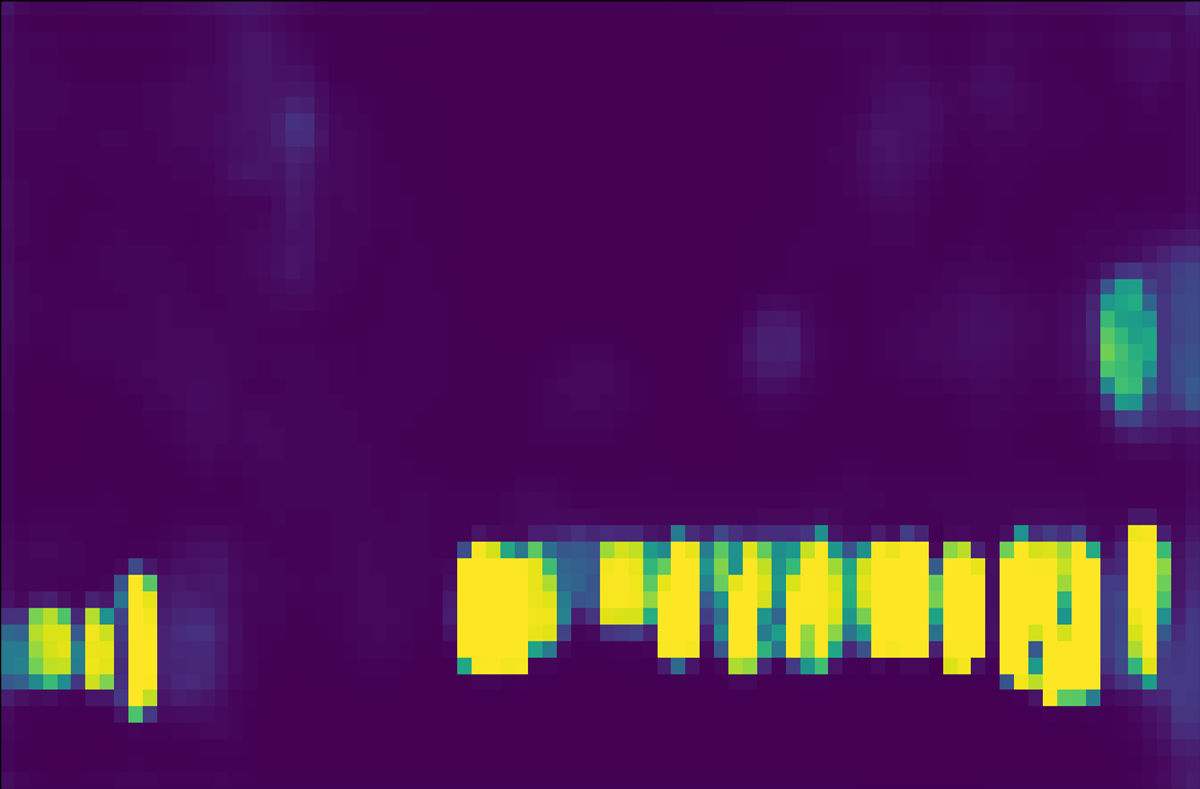}}\vspace{1pt}
    \end{minipage}
    \caption{Score map ($baseline$).}
    \label{fig:scoreml}
  \end{subfigure}
  \hspace{1pt}
  \begin{subfigure}{0.23\textwidth}
    \begin{minipage}{\textwidth}
    \centerline{\includegraphics[width=\textwidth]{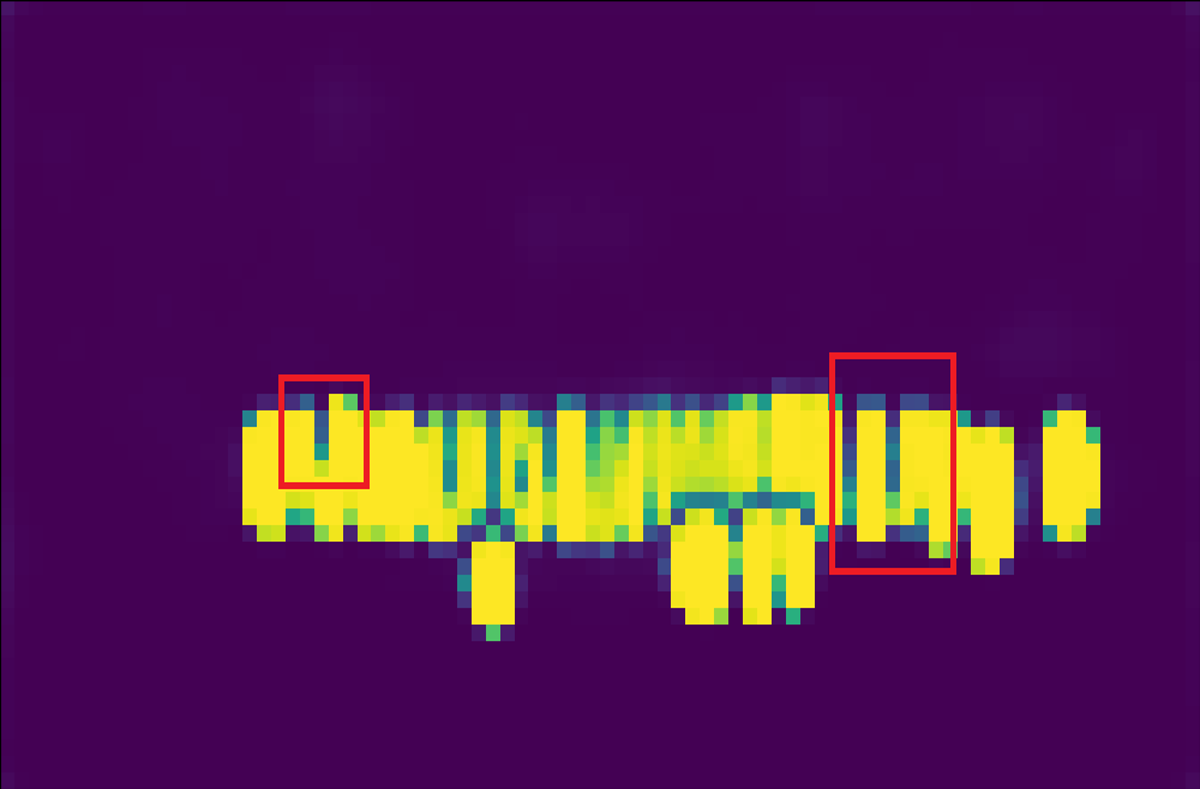}}\vspace{1pt}
    \centerline{\includegraphics[width=\textwidth]{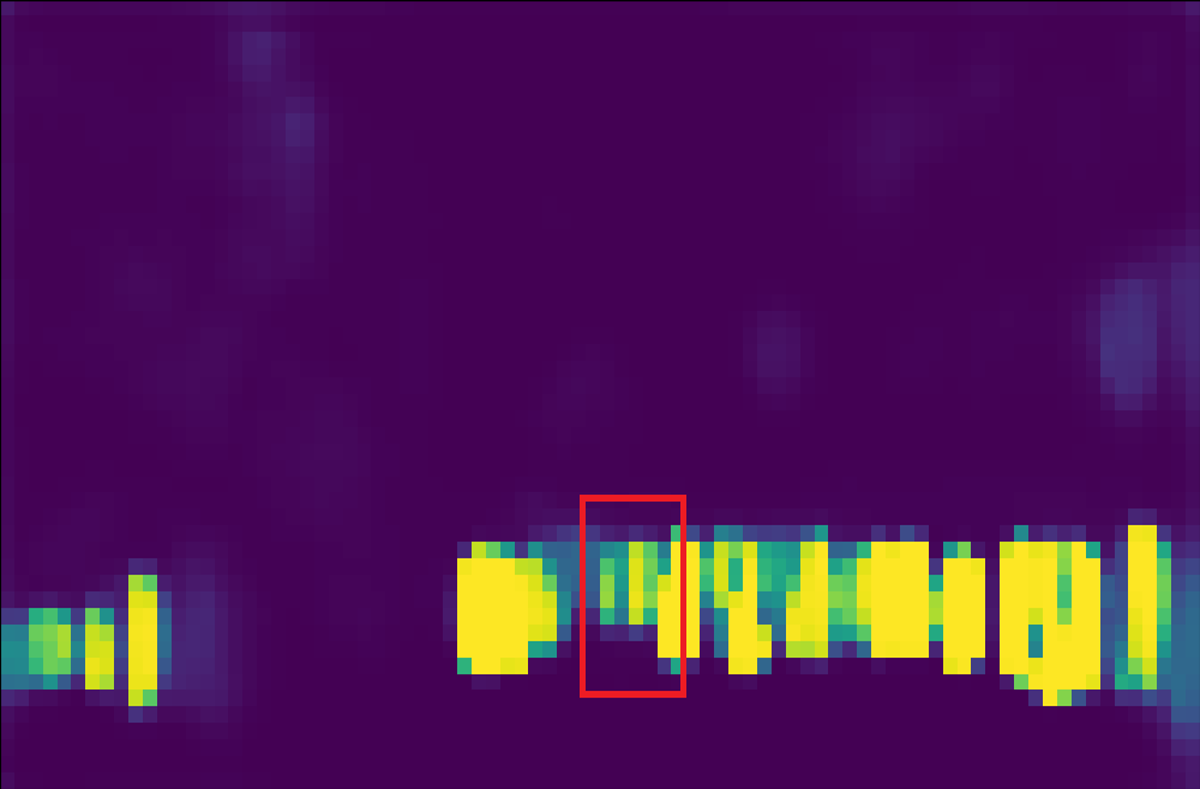}}\vspace{1pt}
    \end{minipage}
    \caption{Score map ($ours$).}
    \label{fig:scoreaevb}
  \end{subfigure}
  \hspace{1pt}
  \begin{subfigure}{0.23\textwidth}
    \begin{minipage}{\textwidth}
    \centerline{\includegraphics[width=\textwidth]{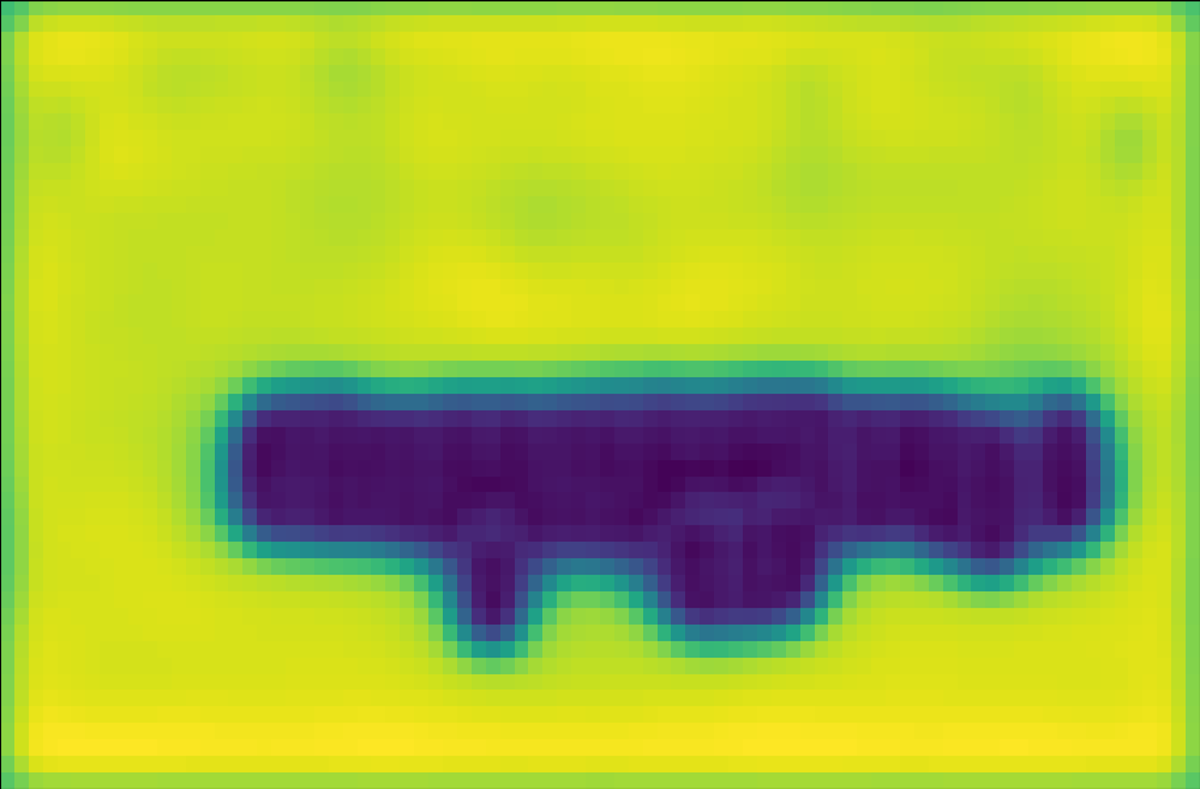}}\vspace{1pt}
    \centerline{\includegraphics[width=\textwidth]{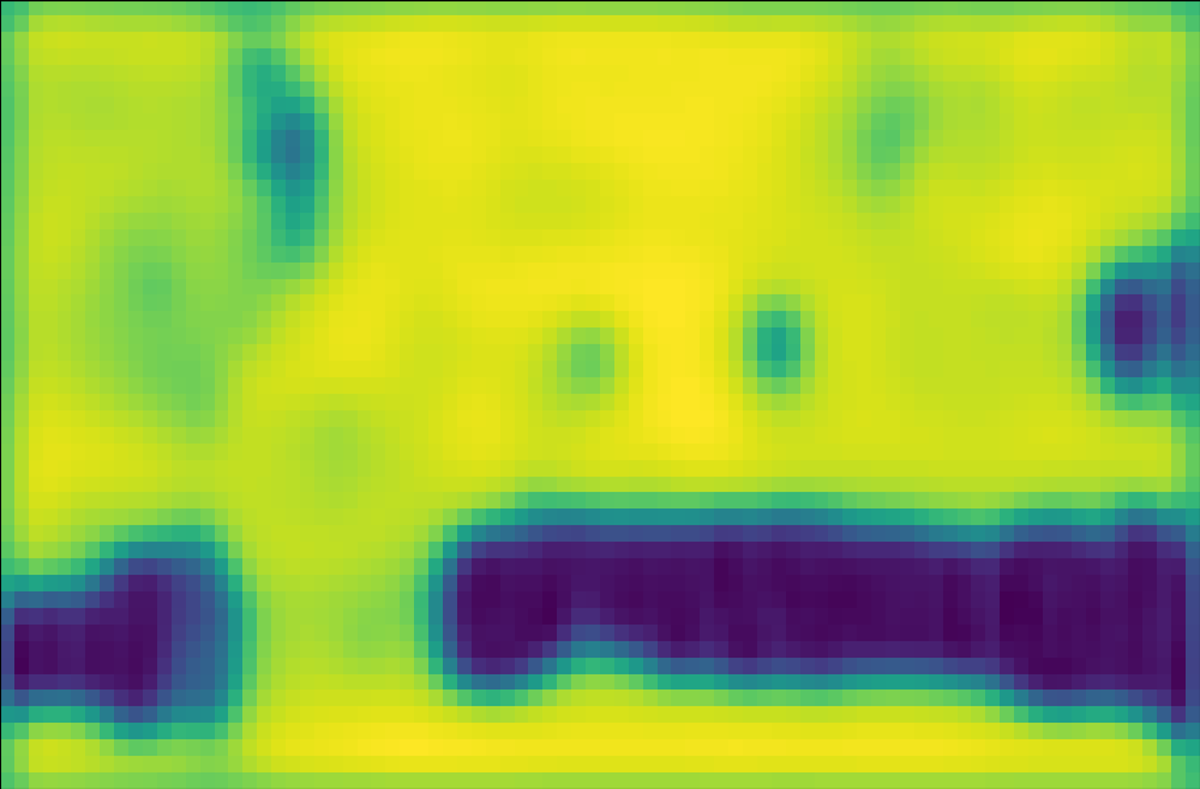}}\vspace{1pt}
    \end{minipage}
    \caption{Variance map ($ours$).}
    \label{fig:varaevb}
  \end{subfigure}
  \caption{\textbf{Score and variance heat map optimized by maximum likelihood (ML) and our method.} These visuals are generated using the feature maps extracted from FPN $P_3$. 
  The variance map is the sum of standard division in log scale: $\sum_{k=1}^4\log\sigma_k$. The red boxes depict successfully detected regions where the ML method failed to detect.}
  \label{fig:visualmap}
  \vspace{-1ex}
\end{figure*}

The final pseudo detection log-likelihood is given by combining \autoref{eq:meanmax} and \autoref{eq:fpl}:
\begin{equation}
  \log \widetilde{p}(\mathbf{x|z}) := w_1\sum_{x_i\in\bf x} \log\mathcal{P}_{i, pos} + w_2\sum_{z_j\in\bf z} \log \mathcal{P}_{j, neg},
\end{equation}
where the factors $w_1 = \frac{0.5}{\|\mathbf{x}_{gt}\|}$, $w_2 = \frac{0.5}{n\|\mathbf{x}_{gt}\|}$ and $\gamma=2$ are kept the same as in \cite{zhang2019freeanchor}.

\section{Experiments} \label{exp}

\subsection{Implementation details}
\textbf{Backbone}.
We adopt the RetinaNet \cite{lin2020focal} architecture with ResNet-50-FPN \cite{lin2017feature, he2016deep} backbone. Following ATSS \cite{zhang2019bridging}, we define one anchor per pixel for simplicity. We present experimental results on FA \cite{zhang2019freeanchor} and FCOS \cite{tian2019fcos}. FA is an anchor-based method of which the four regression variables for the center-point location, width and height are predicted to recover the box. FCOS is an anchor-free method and its regression targets are distances from the location to four sides of the bounding box.

\textbf{Training details}.
We implement the AEVB algorithm in the \emph{mmdetection} \cite{chen2019mmdetection} framework. Pedestrian detection methods are compared on two widely used datasets including the CrowdHuman \cite{shao2018crowdhuman} and the CityPersons \cite{zhang2017citypersons, cordts2016the} datasets. The Crowdhuman consists of 15,000 images with 339,565 human instances for training, and 4,370 images for validation. For CityPersons, we train our models on 2,112 images in the reasonable (R) and highly occluded (HO) subsets, and evaluate on 500 images in the validation set. All experiments are conducted on challenging full-body annotations.

We apply SGD optimizer with an initial learning rate $0.005$, momentum $0.9$, and weight decay $0.0001$. For the CrowdHuman dataset, we train our models on $2$ GTX 1080Ti GPUs with $4$ images per GPU for $24$ epochs and the learning rate is then reduced by an order at the $16^{th}$ and $22^{th}$ epochs. For fair comparison with the RetinaNet baseline on CrowdHuman, multi-scale training and testing are not applied. We resize the short edges of input images to 800 pixels while long edges are kept less than 1333. For CityPersons dataset, we follow the data augmentation setup in Pedestron \cite{hasan2020pedestrian}, applying $2$ images per GPU on $4$ GTX 1080Ti GPUs. The training schedule and learning rate is doubled to that of CrowdHuman setup. The image size for testing is kept the same as the original \emph{i.e.} $2048\times 1024$.

We choose univariate normal distribution as the variational distribution family for the four localization variables. In total, nine variables are predicted for each anchor: one for classification, four for mean $\mu$, and four for standard deviation $\sigma$. The standard deviation is necessary for training but not considered during the inference procedure since the final detection boxes take the maximum likelihood (as for normal distribution) at the means.

\textbf{Evaluation metric}.
We apply log average miss rate (MR) proposed in \cite{dollar2012pedestrian}, which is the average miss rate in log scale over false-positives per image ranging from $[10^{-2}, 10^0]$. A lower miss rate indicates better detection performance.

\subsection{Main result}
\textbf{Comparison to the state-of-the-art}.
On both the CrowdHuman \cite{shao2018crowdhuman} and CityPersons \cite{zhang2017citypersons, cordts2016the} datasets, we compare the AEVB optimized single-stage detectors to
(1) its plain maximum likelihood counterpart (i.e. FreeAnchor baseline);
(2) state-of-the-art general-purpose single-stage object detector RetinaNet \cite{lin2020focal,shao2018crowdhuman} and RFB-Net \cite{liu2018receptive} with offline anchor assignment;
(3) representative two-stage methods: Faster R-CNN \cite{ren2017faster} and Adaptive-NMS \cite{liu2019adaptive}.

We focus on the popular benchmark CrowdHuman for pedestrian detection and most of the comparisons and ablations are conducted on it. Several conclusions can be drawn from \autoref{table:sotaCH}
(1)  Our method is far in excess of the offline learning methods RFB Net and RetinaNet.
(2) The average miss rate of FreeAnchor and FCOS drop from $52.8\%$ to $50.7\%$ and from $48.3\%$ to $47.6\%$ if optimized by our method.
(3) Our optimized two-stage solution Faster R-CNN based on \cite{liu2019adaptive} obtains better miss rate compared to the state-of-the-art Adaptive-NMS method.
(4) We execute our method on a new Faster R-CNN baseline implemented by \cite{chu2020detection} and improve the miss rate up to $40.7\%$ in comparison to OP-MP\cite{chu2020detection} with $41.4\%$ miss rate, which brings the state-of-the-art to a new bar.

We also extend our method to another widely used benchmark CityPersons. Accroding to \autoref{table:sotaCP}, our method outperforms the plain single-stage method RetinaNet, RFB-Net and TLL in crowd detection on CityPersons. The average miss rate of FreeAnchor drops from $14.8\%$ to $13.6\%$ on the CityPersons reasonable subset and from $42.8\%$ to $41.5\%$ on the highly occluded subset by using our method. As for the two-stage detectors, we verify the superiority of our method over other two-stage methods including Repp Loss \cite{wang2017repulsion} and Adaptive-NMS \cite{liu2019adaptive}. These results indicate that learning-to-match with the AEVB algorithm outperforms plain maximum likelihood methods on both datasets.

\begin{table}[t!]
  \centering
  \small
  \caption{Comparison vs. other methods on the CrowdHuman dataset. Lower MR is better.}
  \vspace{-1ex}
  \label{table:sotaCH}
  \begin{tabular}{@{}lll@{}}
    \toprule
    & Method                    & MR $\downarrow$ \\ \midrule
    \multirow{6}{*}{Single-stage}
    & RFB Net \cite{liu2018receptive, liu2019adaptive}          & 65.2             \\
    & RetinaNet \cite{lin2020focal, shao2018crowdhuman}         & 63.3             \\
    & FreeAnchor ($baseline$)                                   & 52.8             \\
    & FCOS ($baseline$)                                         & 48.3             \\
    & FreeAnchor ($ours$)                                       & \textbf{50.7}    \\
    & FCOS ($ours$)                                             & \textbf{47.7}    \\ \midrule
    \multirow{8}{*}{Two-stage}
    & Faster R-CNN ($impl.\ by\ \cite{liu2019adaptive}$)        & 52.4              \\
    & Faster R-CNN ($our\ impl.$)                               & 51.2             \\
    & Adaptive NMS \cite{liu2019adaptive}                       & 49.7              \\
    & Faster R-CNN ($ours$)                                     & \textbf{48.8}    \\ \cline{2-3}
    & Faster R-CNN ($impl.\ by\ \cite{chu2020detection}$)       & 42.9              \\
    & Faster R-CNN ($our\ impl.$)                               & 42.4             \\
    & GossipNet \cite{hosang2017learning}                       & 49.4              \\
    & RelationNet \cite{hu2018relation}                         & 48.2              \\
    & OP-MP \cite{chu2020detection}                             & 41.4              \\
    & Faster R-CNN ($ours$)                                     & \textbf{40.7}    \\
    \bottomrule
  \end{tabular}
\end{table}

\begin{table}[t!]
  \centering
  \small
  \caption{Result on CityPersons tested at $2048\times1024$. R: reasonable subset; HO: highly occluded subset. Performance evaluated with Miss Rate (lower is better).}
  \vspace{-1ex}
  \label{table:sotaCP}
  \begin{tabular}{@{}llll@{}}
    \toprule
            & Method                                             & R              & HO              \\ \midrule
    \multirow{5}{*}{Single-stage}
            & RetinaNet \cite{lin2020focal, hasan2020pedestrian} & 15.6           & 49.9            \\
            & FreeAnchor                                         & 14.8           & 42.8           \\
            & TLL \cite{song2018small}                           & 14.4           & 52.0*           \\
            & RFB Net \cite{liu2018receptive, liu2019adaptive}   & 13.9           & -               \\
            & FreeAnchor($ours$)                                 & \textbf{13.6} & \textbf{41.5}  \\\midrule
    \multirow{4}{*}{Two-stage}
            & Faster R-CNN \cite{wang2017repulsion}              & 14.6           & 60.6*           \\
            & Rep Loss \cite{wang2017repulsion}                  & 13.2           & 56.9*           \\
            & Adaptive-NMS \cite{liu2019adaptive}                & 12.9           & 56.4*           \\
            & Faster R-CNN ($ours$)                              & \textbf{12.7} & \textbf{54.6}*  \\
    \bottomrule
  \end{tabular}
  \\{\footnotesize * denotes the detector is not trained on the HO subset.}
\end{table}

\textbf{Relation between score and variance}.
We show the score map and variance map learned by the AEVB algorithm in comparison to the maximum likelihood method in \autoref{fig:visualmap}. For fairness, single anchor design and IoU likelihood are applied in both methods. The score map learned by AEVB (\autoref{fig:scoreaevb}) shows a more compact assignment and a cleaner background compared to the score map optimized by ML (\autoref{fig:scoreml}). Furthermore, we plot the variance map in \autoref{fig:varaevb} as 
the sum of log standard deviation, i.e., $\log\sigma_{dx} + \log\sigma_{dy} + \log\sigma_{\log w} + \log\sigma_{\log h}$. Intuitively, the variance is closer to one for background pixels due to KL regularization (1st term in \autoref{eq:elbo}) while it is significantly lower for foreground pixels due to the data term (2nd term in \autoref{eq:elbo}).

\begin{figure}
  \centering
  \begin{subfigure}{0.23\textwidth}
    \centering
    \includegraphics[width=\textwidth, clip]{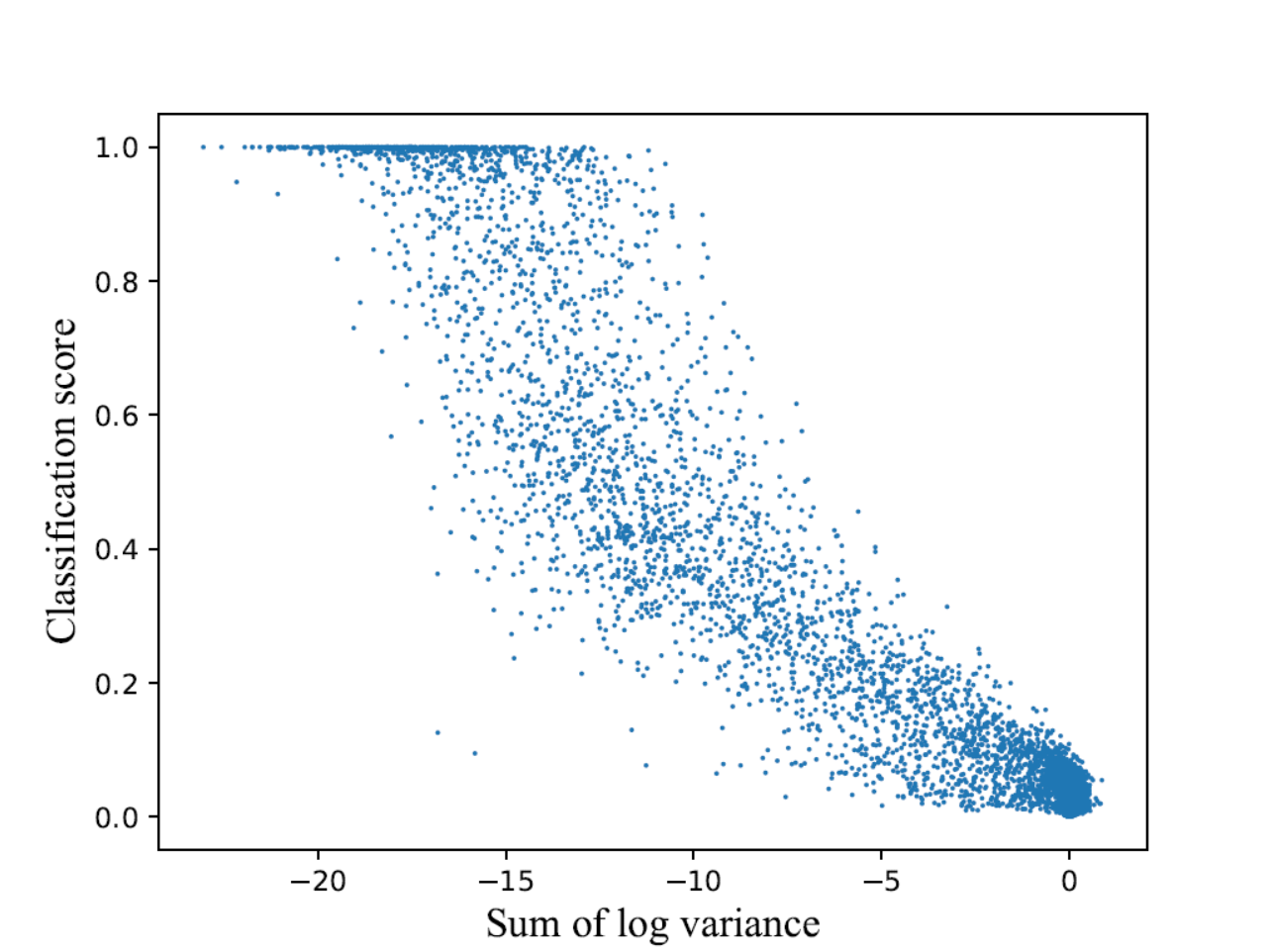}
    \captionsetup{font={scriptsize}}
    \caption{Classification score vs. Variance}
    \label{fig:var_score}
  \end{subfigure}
  \begin{subfigure}{0.23\textwidth}
    \centering
    \includegraphics[width=\textwidth, clip]{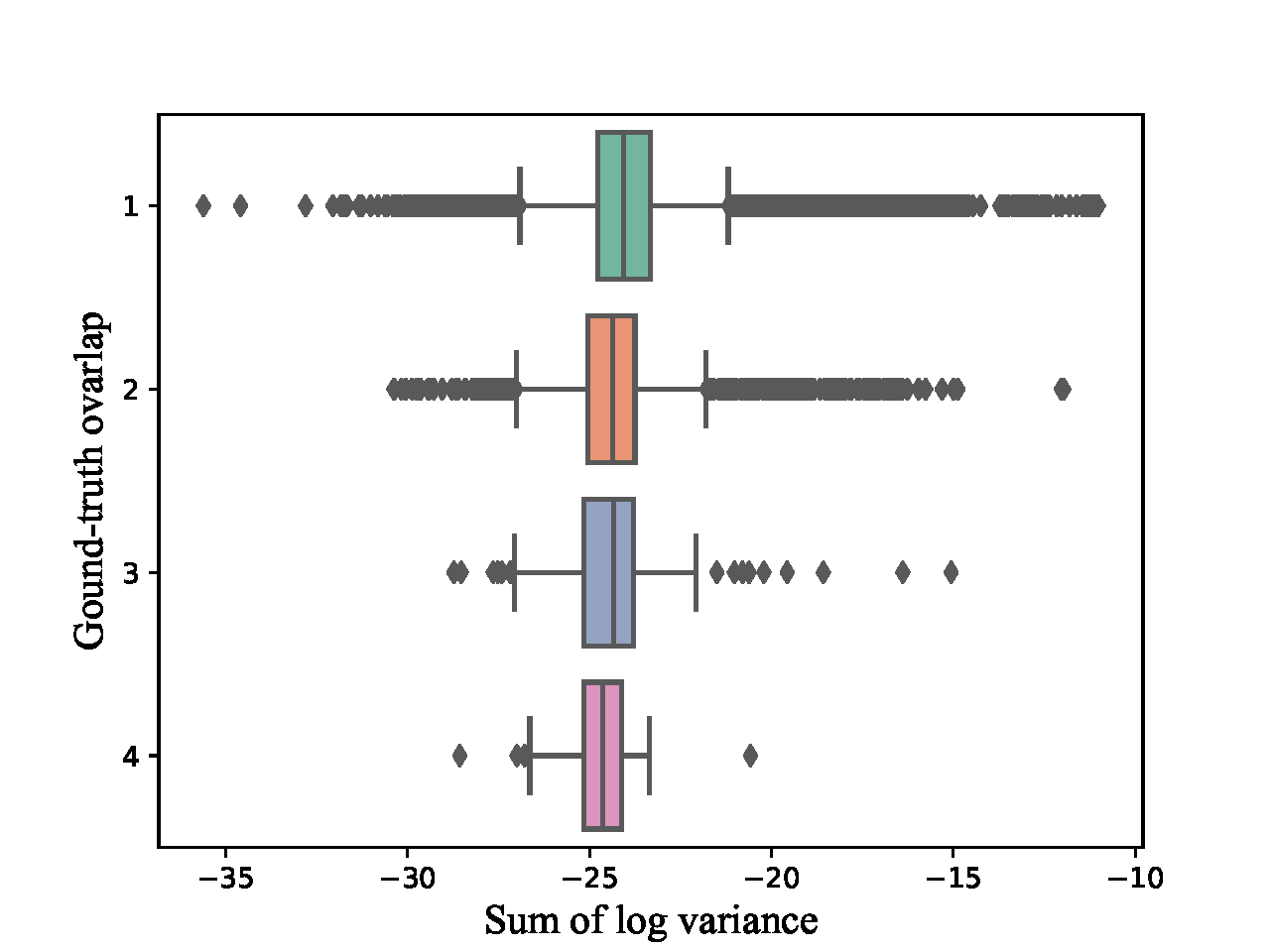}
    \captionsetup{font={scriptsize}}
    \caption{Ground-truth overlap vs. Variance}
    \label{fig:var_ovlp}
  \end{subfigure}
  \caption{\textbf{The correlation between the variance of proposals with the ground-truth overlap and the classification confidence.}  The ground-truth overlap in (b) is represented by the number of matched ground-truth boxes that have IoU overlap of larger than 0.5 with the corresponding proposals. The variance in (a) and (b) is in log scale.}
  \label{fig:varrelation}
  \vspace{-2ex}
\end{figure}

We also visualize the predicted variance of dense proposals versus their classification score and overlap with groundtruth respectively. In \autoref{fig:var_score}, proposals with low confidence ususally yield large variance to encourage a broader searching space for potential matched ground-truth boxes, while most confident proposals keep low variance to ensure stable regression. \autoref{fig:var_ovlp} shows that the median of proposal variance and the number of proposals with large variance outlier decrease as the number of matched ground-truth boxes increases. Thus, our method can be termed as occlusion-aware. These phenomenons not only help to stabilize the optimization in dense crowd region, but also improve the miss rate.

\subsection{Ablation study}
\textbf{Detection pipelines.}
As analyzed in \autoref{plhd}, our method can smooth the IoU gradient map, facilitating a more feasible gradient descent. We verify our hypothesis in \autoref{table:modify}. It shows that the effect of using IoU loss is quite notable (1.22\%) and adding AEVB to detectors based on IoU loss further enhances the detection performance (0.94\%). Likewise, FCOS with our approach improves MR by 0.74\% over its baseline. Based on these results, we can conclude that both the variational dense proposal and IOU-based detection extraction module jointly improve the performance over occluded scenes. Moreover, our method can generalize well to both anchor-based and anchor-free methods.

\begin{table}[t!]
  \centering
  \small
  \caption{Ablation experiments on CrowdHuman with different box models. Baseline models
  are compared to our method. $modified$: replace L1 loss of location targets by IoU loss. $ours$: modify detector by replacing with IoU loss and optimize it by our AEVB algorithm. Performance evaluated with Miss Rate (lower is better).}
  \vspace{-1ex}
  \label{table:modify}
  \begin{tabular}{@{}l|cc|c@{}}
    \toprule
    Method         & IoU loss        & AEVB           & MR $\downarrow$   \\ \midrule
    FA ($baseline$)  &                 &                & 52.83             \\
    FA ($modified$)  & \checkmark      &                & 51.61             \\
    FA ($ours$)      & \checkmark      & \checkmark     & \textbf{50.67}    \\ \midrule
    FCOS ($baseline$)& \checkmark      &                & 48.31             \\
    FCOS ($ours$)    & \checkmark      & \checkmark     & \textbf{47.57}    \\
    \bottomrule
  \end{tabular}
\end{table}

\textbf{The KL factor}.
The scaling factor $\alpha$ can be tuned to match the pseudo detection likelihood $\widetilde{p}(\bf x|z)$. We tested our algorithm on the CrowdHuman \cite{shao2018crowdhuman} dataset with scaling factors $\alpha$ ranging from $0$ to $10^{-1}$. Results in \autoref{table:factor} indicate that $10^{-2}$ yields the best performance with our algorithm outperforming the plain maximum likelihood method (FreeAnchor + IoU) on the CrowdHuman dataset by 0.94\% MR. Note that this value is also a stable choice for the CityPerson dataset. The best values of KL factor for FCOS and Faster R-CNN are nearly the same ($10^{-3}$) which implies that the optimal setting of $\alpha$ is not sensitive to concrete box models. 

\begin{table}
  \centering
  \small
  \caption{Ablation experiments evaluated on CrowdHuman dataset for the KL factor. Lower MR is better.}
  \vspace{-1ex}
  \label{table:factor}
  \begin{tabular}{@{}l|ccc@{}}
    \toprule
    KL factor $\alpha$      & $10^{-4}$     & $10^{-3}$   & $10^{-2}$  \\ \midrule
    FreeAnchor($ours$)      & 51.96     & 51.27    & \textbf{50.67}  \\
    FCOS($ours$)            & 47.78     & \textbf{47.57}   & 48.69  \\
    Faster R-CNN($ours$)    & 40.93    & \textbf{40.69}  & 41.19   \\
    \bottomrule
  \end{tabular}
  \vspace{-1.2ex}
\end{table}

\textbf{The number of anchors}.
The results in \autoref{table:anchors} shows that the effect of increasing the number of anchors sample is non-negligible. FreeAnchor with two-anchor design can enhance the MR by 0.36\% while four-anchor design further improve it by 0.38\%. FCOS with four-anchor design can achieve a 0.08\% improvement. The training time consumption of both methods increase by about 8\% compared to the baselines. And their multiple-anchor designs only incur acceptable addition of training runtime. The testing runtime is not affected as we directly take the mean of each variable as the final localization results, following max-likelihood principle.
Qualitatively, single-anchor design are sufficient for good performance while being more efficient with multiple-anchor design. However, considering simpler design and faster training, we keep to the single anchor design for all other reporting.

\begin{table}
  \centering
  \small
  \caption{Ablation experiments evaluated on CrowdHuman with varying number of anchors per pixel.}
  \vspace{-1ex}
  \label{table:anchors}
  \begin{tabular}{@{}lccc@{}}
    \toprule
    Method                  & anchor(s) $n$        & MR $\downarrow$  & training time        \\ \midrule
    FreeAnchor($baseline$)  & $1$                  & 52.83            & 8.06 h                   \\ \midrule
    \multirow{3}{*}{FreeAnchor($ours$)}
                            & $1$                  & 50.67            & 8.63 h                   \\
                            & $2$                  & 50.31            & 9.77 h                   \\
                            & $4$                  & \textbf{49.93}   & 11.91 h                  \\ \midrule
    \multirow{2}{*}{FCOS($ours$)}
                            & $1$                  & 47.57            & 10.88 h                  \\
                            & $4$                  & \textbf{47.49}   & 12.53 h                  \\
    \bottomrule
  \end{tabular}
  \vspace{-1.2ex}
\end{table}

\textbf{Levels of occlusion}.
The human instances in CrowdHuman dataset are split into 3 subsets according to the level of occlusion, \emph{i.e.}, the maximum IoU to other human instances. \autoref{table:occlu} shows that the AEVB algorithm outperforms the offline method RetinaNet and online method FreeAnchor across all levels of occlusion especially on the \textit{Partial} subset where the intra-class occlusion is still reasonably high
The overall detection performance on the \textit{Heavy} occlusion subset is expectantly poorer than the other subsets, but improvements are noticeable as well with AEVB. 

\begin{table}
  \centering
  \small
  \centering
  \caption{Detection performance evaluated on different levels of occlusion on the CrowdHuman dataset. Performance evaluated in Miss Rate (lower is better).}
  \vspace{-1ex}
  \label{table:occlu}
  \begin{tabular}{@{}c|cccc@{}}
    \toprule
    Occlusion        & Bare           & Partial        & Heavy          & All            \\
    IoU              & $[0, 0.3]$     & $(0.3, 0.7]$   & $(0.7, 1]$     & $[0, 1]$       \\
    \midrule
    \# instances     & 47469          & 49146          & 2866           & 99481          \\
    \midrule
    RetinaNet        & 54.50          & 59.39          & 65.69          & 59.97          \\
    FA($baseline$)   & 49.32          & 57.15          & 66.48          & 52.83          \\
    FA($modified$)   & 47.83          & 52.97          & 65.15          & 51.61          \\
    FA($ours$)       & \textbf{46.91} & \textbf{52.19} & \textbf{64.32} & \textbf{50.67} \\
    \bottomrule
  \end{tabular}
  \vspace{-1ex}
\end{table}

\subsection{Extension to two-stage methods}
Although we mainly focus on single-stage pedestrian detection, the proposed optimization algorithm can be flexibly extended to two-stage methods. We applied the proposed AEVB algorithm to optimize the RPN in Faster R-CNN \cite{ren2017faster} while leaving the network structure and the second stage unchanged. \autoref{table:sotaCH} and \autoref{table:sotaCP} list the experimental results on both evaluated datasets, which show Faster R-CNN + AEVB clearly outperforms the original Faster R-CNN and its extensions \cite{chu2020detection, hosang2017learning, hu2018relation, liu2019adaptive}. This verifies the generalization capability of the proposed AEVB algorithm.

\section{Conclusion}
We reformulate single-stage pedestrian detection as a \textit{variational inference} problem and propose a customized Auto-Encoding Variational Bayes (AEVB) algorithm to optimize the problem. For the determination of intractable detection likelihood, we provide a relaxed solution which works well on both FreeAnchor and FCOS box models. We demonstrate the potential of this formulation to propel pedestrian detection performance of single-stage detectors to higher level, while showing that the proposed optimization can also be generalized to two-stage detectors.

\noindent \textbf{Acknowledgement.} The paper is supported in part by the following grants: China Major Project for New Generation of AI Grant (No.2018AAA0100400), National Natural Science Foundation of China (No. 61971277).

  {\small
    \bibliographystyle{ieee_fullname}
    \bibliography{references}
  }

\end{document}